\documentclass[11pt]{article}

\usepackage[preprint]{acl}

\usepackage{times}
\usepackage{latexsym}

\usepackage[T1]{fontenc}

\usepackage[utf8]{inputenc}

\usepackage{microtype}

\usepackage{inconsolata}

\usepackage{graphicx}
\usepackage{todonotes}
\usepackage{comment}
\usepackage{amsmath}
\usepackage{booktabs}
\usepackage{amssymb}
\usepackage{xcolor}
\usepackage{tikz}
\newcommand{\circled}[1]{%
  \tikz[baseline=(char.base)]{
    \node[shape=circle,draw,inner sep=1pt] (char) {#1};
  }%
}
\usepackage{xcolor}
\usepackage{xspace}

\definecolor{r1}{HTML}{0072B2}
\definecolor{r2}{HTML}{D55E00}
\definecolor{r3}{HTML}{008C95}   
\definecolor{r3}{HTML}{009E73}
\definecolor{r4}{HTML}{CC79A7}
\definecolor{r5}{HTML}{E69F00}
\definecolor{withinset}{HTML}{5B9BD5}   
\definecolor{overall}{HTML}{168A16}  
\definecolor{nolit}{HTML}{5B00B5}   

\newcommand{\crone}{\textcolor{r1}{\texttt{R1}}\xspace}
\newcommand{\crtwo}{\textcolor{r2}{\texttt{R2}}\xspace}
\newcommand{\crthree}{\textcolor{r3}{\texttt{R3}}\xspace}
\newcommand{\crfour}{\textcolor{r4}{\texttt{R4}}\xspace}
\newcommand{\crfive}{\textcolor{r5}{\texttt{R5}}\xspace}
\newcommand{\cnolit}{\textcolor{nolit}{\textit{No Literature}}\xspace}

\newcommand{\cwithin}{\textcolor{withinset}{\textit{within-set}}\xspace}
\newcommand{\coverall}{\textcolor{overall}{\textit{overall}}\xspace}

\title{Tracing the Influence of Retrieved Literature on Scientific Ideation in LLMs}

\author{
\textbf{Zhipin Wang$^{1}$, Koel Dutta Chowdhury$^{1}$, Christoph Leiter$^{1}$, Steffen Eger$^{1}$}\\
University of Technology Nuremberg (UTN), Germany\\
\texttt{zhipin.wang@utn.de, steffen.eger@utn.de} \\
$^1$\texttt{NLLG} (\url{https://nl2g.github.io/})
}

\begin{document}
\maketitle
\begin{abstract}

Large language models (LLMs) are increasingly used to generate research ideas from scientific literature. 
Existing systems combine literature retrieval with search, generation, and refinement, making it difficult to determine how the retrieval strategy itself shapes the ideas generated.
We isolate this contribution by comparing five retrieval strategies and a No Literature baseline under a shared generation protocol, using 45,000 ideas across 15 machine learning topics and ten open-weight models.
Relative to no literature basline, all five retrieval strategies reduce diversity among ideas generated from the same set of papers by $5.6\%$--$17.9\%$, while most increase diversity by $10.2\%$--$18.1\%$ when ideas from different sets are pooled. Effects on idea quality and novelty are modest on average but vary across topics.
We further apply a JSD-based attribution method to trace the influence of individual input papers on each idea's motivation, core idea, approach, and validation. In $79.3\%$ of ideas, only one or two distinct papers receive the highest attribution scores across these four parts. These leading sources often recur across ideas generated from the same set of papers, and groups with more varied source attribution contain more diverse ideas ($r=0.70$). 
These findings motivate various sets of literature and the use of attribution to guide the choice of sources for subsequent ideas.

\end{abstract}

\section{Introduction}
\label{sec:introduction}

\begin{figure}[t]
    \centering
    \includegraphics[width=0.93\columnwidth]{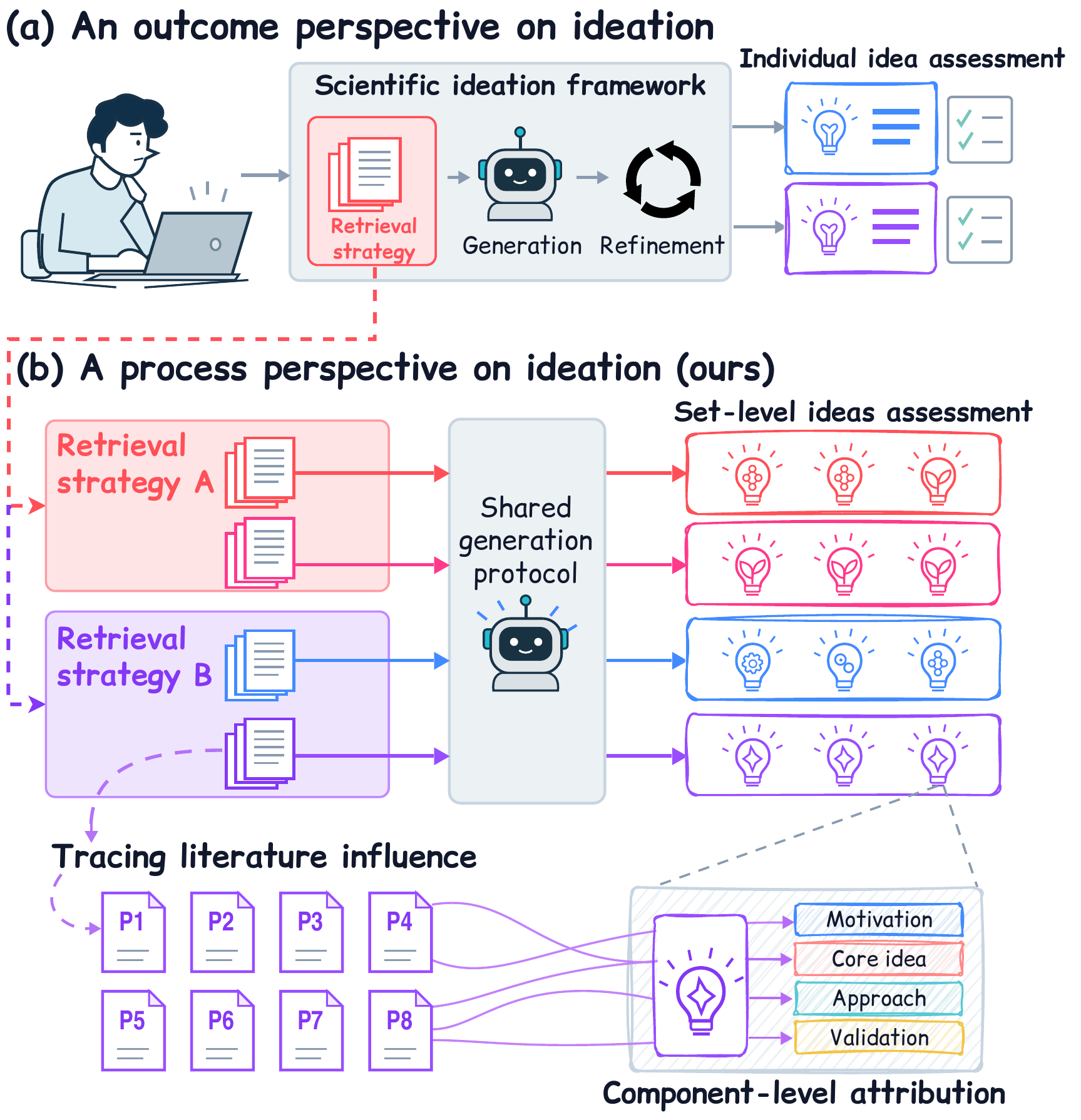}
\caption{
\textbf{How do retrieval strategies shape scientific exploration?}
(a) Assessing individual ideas does not reveal how retrieval shapes the range of ideas explored.
(b) We compare retrieval strategies under a shared generation protocol, assess idea sets, and trace literature influence on idea components.
}
    \label{fig:conceptual_gap}
\end{figure}

Scientific ideation requires creativity, commonly defined in terms of originality and effectiveness~\citep{runco2012standard}. Reflecting these requirements, research on LLM-based scientific ideation commonly evaluates individual ideas in terms of novelty, feasibility, and scientific value~\citep{si2024novel,eger2026thaisci,chen2026ideationarena}. A dynamic perspective on creativity further emphasizes potential originality and effectiveness and considers the entire ensemble of products generated by a creative process~\citep{corazza2016dynamic}. Recent work also incorporates diversity into LLM creativity evaluation and explicitly models and optimizes scientific ideation workflows~\citep{sen-etal-2026-automated}. Evaluating an ideation system therefore also requires considering the range of research directions produced across generations (Figure~\ref{fig:conceptual_gap}).

Literature retrieval is one of the key steps in this process, determining which papers a model receives during ideation. Existing systems combine retrieval with their own generation, search, and refinement procedures~\citep{wang2024scipip,baek2025researchagent,xiong-etal-2026-evosci,wang2026flowpie}. 
However, this integration makes it difficult to isolate the effect of literature selection from other stages of the ideation process. This distinction matters because different sets of papers may steer the model toward different research directions, even when the resulting sets of ideas have similar average quality or novelty. In particular, it remains unclear whether changing the supplied papers produces more varied ideas than generating multiple ideas from the same papers.

To address this question, we isolate literature selection under a shared generation protocol (Figure~\ref{fig:conceptual_gap}). For each research topic, each retrieval strategy provides multiple sets of papers, and we generate several ideas using each set as context. Following recent work~\citep{DBLP:journals/tmlr/HouZLBBLJBCCKL26}, we evaluate generated ideas along three complementary dimensions: \textit{quality}, \textit{novelty}, and \textit{diversity}. Quality and novelty characterize individual ideas, whereas diversity captures the range of directions represented by a set of ideas. 
We compare diversity among ideas generated from the same papers with diversity among all ideas generated from different sets of papers.

We complement this set-level analysis with context attribution based on input erasure~\citep{li2017understandingneuralnetworksrepresentation}, adapting methods developed for retrieval-augmented generation~\citep{cohenwang2024contextcite,li2025arcjsd} to scientific ideation. We remove retrieved papers one at a time and measure changes in the model's predictions for the same generated idea.

Our study\footnote{We will release our code and data upon acceptance.} examines \textbf{45,000 research ideas} across five retrieval strategies
and a \cnolit baseline, 15 machine learning topics, and ten open-weight models. Our main findings are:

\vspace{0.2cm}
    \circled{1}~\textbf{Retrieval narrows diversity among ideas generated
    from the same papers but can broaden it across different paper selections.}
    Relative to \cnolit, all five retrieval strategies reduce \cwithin
    diversity by $5.6\%$--$17.9\%$, while four increase diversity by
    $10.2\%$--$18.1\%$ when ideas generated from different literature sets are considered together. Differences in mean quality and novelty are comparatively modest.

\vspace{0.2cm}

\circled{2}~\textbf{Dependence on retrieved literature is concentrated and
    recurrent.} Our attribution analysis identifies at most two of the eight retrieved
papers as leading sources for $79.3\%$ of ideas, with a single leading
source for $39.5\%$.
    Among ideas generated from the same papers, the leading paper for a given component is shared by $38.7\%$--$43.2\%$ of idea pairs. Groups with more varied core-idea source dependence also contain more diverse ideas
    ($r=0.70$).

\vspace{0.2cm}

\circled{3}~\textbf{Retrieval's average diversity advantage shrinks
across model updates, while its quality advantage grows.}
    Across matched model updates, retrieval's average diversity advantage
    decreases by approximately $0.030$, while its relative quality advantage
    increases by approximately $0.249$. These changes differ substantially
    across model families.

\section{Idea Generation and Attribution}
\label{sec:ideation_protocol}
\label{sec:framework}

We separate variation associated with retrieved literature from variation in ideas generated under the same context and variation across models. For the same research topic, we compare (i) ideas generated from one set of papers, (ii) ideas generated from different sets under the same model, and (iii) ideas generated from the same set across models. These comparisons follow a shared generation protocol across retrieval strategies.

\subsection{Idea Generation Protocol}

Let $\mathcal{P}$ denote the \emph{paper pool}, the set of papers
available to the retrieval strategies. Given a research topic $q$,
a retrieval strategy $R$ constructs a \emph{literature set}
$\mathcal{D} \subseteq \mathcal{P}$, a set of papers supplied together
as context for idea generation:
\begin{equation}
    \mathcal{D} \sim R(\cdot \mid q,\mathcal{P}).
    \label{eq:retrieval}
\end{equation}

A model $\theta$ receives the research topic $q$ and literature set
$\mathcal{D}$ through a common instruction template $I$ and generates
a research idea $y$:
\begin{equation}
    y \sim p_\theta\!\left(\cdot \mid I(q,\mathcal{D})\right).
    \label{eq:ideation}
\end{equation}

Each idea $y$ consists of a title and four components:
\emph{motivation}, which states the research problem and why it matters;
\emph{core idea}, which states the central contribution;
\emph{approach}, which describes how the idea would be implemented;
and \emph{validation plan}, which outlines how it would be evaluated.

For each literature set $\mathcal{D}$, we independently generate
several ideas using the same model and refer to them as a
\emph{generation group}. Because the topic, papers, instruction,
and model are fixed within a group, differences reflect variation
among ideas generated under the same conditions. We compare these
groups across literature sets, retrieval strategies, and models.

\subsection{Retrieval Strategies}
\label{sec:retrieval}

For each topic $q$, a retrieval strategy selects a literature set $\mathcal{D}$ of
eight papers from the shared corpus. We compare five strategies that differ in how
they use semantic relevance and citation structure:

\begin{enumerate}
    \item[\textbf{\crone}] \textbf{Semantic Top-$K$.}
    Papers are ranked by semantic similarity to $q$, following semantic retrieval
    used in scientific ideation~\citep{wang2024scimon}. We sample eight papers from
    the top 100 candidates to obtain different literature sets.

    \item[\textbf{\crtwo}] \textbf{Most-Cited Top-$K$.}
    We prioritize highly cited papers among topic-relevant candidates, following
    citation-based selection in ResearchAgent~\citep{baek2025researchagent}, and
    sample eight papers from the top 100.

    \item[\textbf{\crthree}] \textbf{Semantic + Entity + Co-citation.}
    We adapt SciPIP~\citep{wang2024scipip}, combining semantic similarity with
    scientific entities and co-citation relations.

    \item[\textbf{\crfour}] \textbf{Core-Paper Reference Expansion.}
    We select a core paper and construct $\mathcal{D}$ from the paper and its
    references, following ResearchAgent~\citep{baek2025researchagent}.

    \item[\textbf{\crfive}] \textbf{Chain of Ideas.}
    We follow backward references and forward citations to construct a sequence
    of related papers across time~\citep{li2024chain}.
\end{enumerate}

\noindent
\textbf{\cnolit.}
As a baseline, we set $\mathcal{D}$$=\varnothing$ and provide only the research topic.

\begin{figure}[t]
    \includegraphics[width=\columnwidth]{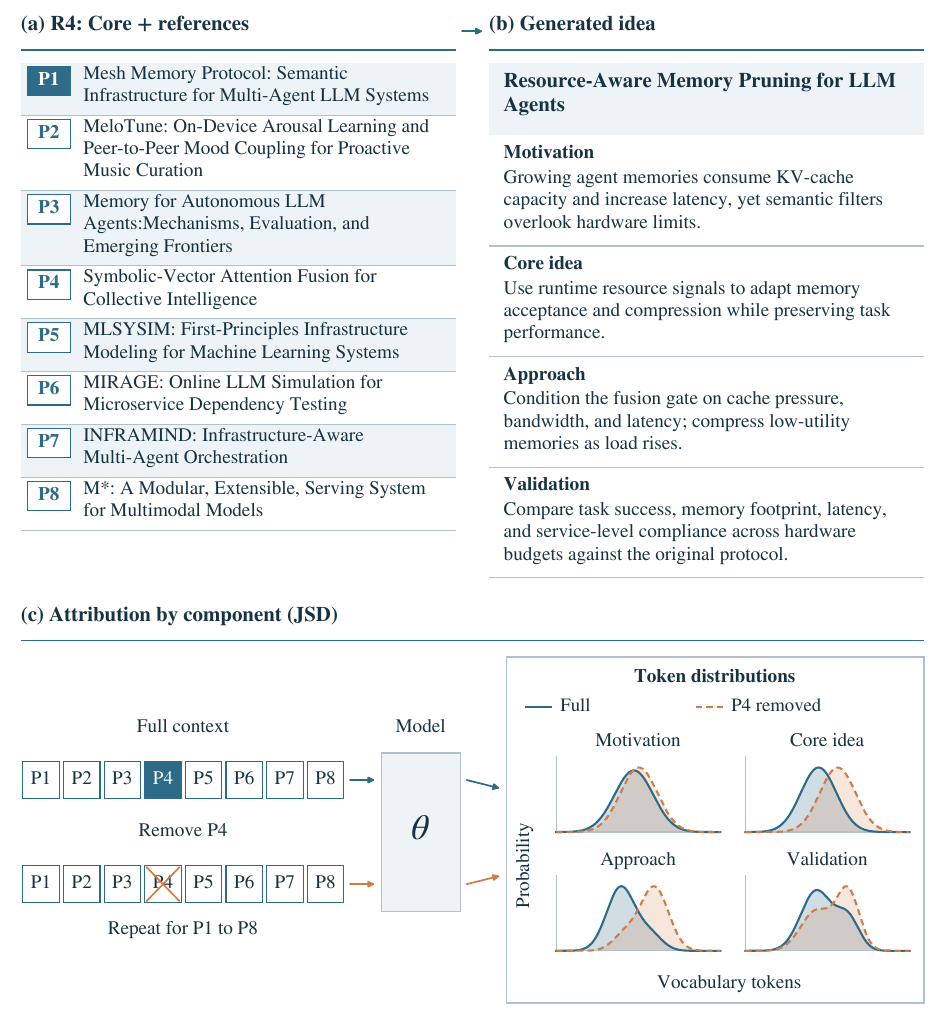}
    \caption{\textbf{An example of generated idea and attribution computation.}
    (a) \crfour retrieves a core paper (P1) and its references; (b) The generated idea is shown through condensed component text.
    (c) Four pairs of curves illustrate distribution changes for the four components under leave-one-paper-out scoring of the generated idea.}
    \label{fig:ideation_example}
\end{figure}

\subsection{Component-Level Attribution}
\label{sec:attribution_method}
\label{sec:attributio}

Finally, we examine how individual retrieved papers shape different
components of a generated idea. Following context-ablation
approaches~\citep{cohenwang2024contextcite,li2025arcjsd}, we remove one
retrieved paper at a time and rescore the same generated text, keeping its
preceding tokens fixed.

Let $\mathcal{D}$ denote the retrieved literature set and
$\mathcal{D}_{-k}$ the same set with paper $d_k$ removed. At token position
$j$, with generated prefix $z_{<j}$ held fixed, we compare the next-token
distributions
\begin{equation}
\begin{aligned}
\pi_j &=
p_\theta\!\left(\cdot \mid I(q,\mathcal{D}), z_{<j}\right), \\
\pi_j^{-k} &=
p_\theta\!\left(\cdot \mid I(q,\mathcal{D}_{-k}), z_{<j}\right).
\end{aligned}
\label{eq:conditional_distributions}
\end{equation}

We compute paper-level attribution scores separately for the motivation, core idea, approach, and validation components. Let $\mathcal{T}_c$ denote the token positions
belonging to component $c$. The score for paper $d_k$ and component $c$ is
\begin{equation}
A_{k,c}
=
\frac{1}{|\mathcal{T}_c|}
\sum_{j\in\mathcal{T}_c}
\operatorname{JS}\!\left(\pi_j,\pi_j^{-k}\right),
\label{eq:component_jsd}
\end{equation}
where $\operatorname{JS}$ denotes Jensen--Shannon divergence. 
Higher scores indicate that removing \(d_k\) induces larger changes in the model's next-token distributions, suggesting a stronger influence on the corresponding idea component.

For comparisons across papers, we normalize the scores within each component,
\begin{equation}
w_{k,c}
=
\frac{A_{k,c}}
{\sum_{\ell=1}^{|\mathcal{D}|} A_{\ell,c}},
\label{eq:normalized_attribution}
\end{equation}
and define the paper with the largest $w_{k,c}$ as the \emph{leading paper}
for that component. We compare these distributions across components and
across ideas generated from the same literature set to measure source
concentration, source recurrence, and variation in source dependence. Figure~\ref{fig:ideation_example} shows an example of a generated idea.

\section{Experimental Setup}
\label{sec:experimental_setup}
We now instantiate our study design with concrete topics, evaluated models, and evaluation metrics. Full evaluation configurations, sampling procedures, and statistical protocols are provided in Appendix~\ref{app:exp_protocol}.

\subsection{Topics and Literature}

We study 15 machine learning topics derived from the primary areas of ICLR 2026,
merging closely related areas. We collect papers first submitted to arXiv between
January 1 and July 23, 2026. For each topic $q$, we rank papers by semantic relevance
and retain the top 2,000. After deduplication across topics, the shared corpus contains
24,836 papers. Retrieval uses each paper's title and abstract.

\subsection{Models and Generation}
\label{sec:models}

We evaluate ten open-weight models from four families, using the 2025 models Qwen3-2507 (30B), Gemma 3 (27B), Nemotron 3 Nano (30B), and GLM-4 (32B) for the main retrieval comparison. Temporal comparisons with the corresponding 2026 releases and the 2024 Qwen and Gemma checkpoints are reported in the Appendix~\ref{app:temporal}.
Open weights enable access to the next-token distributions required for paper-deletion analysis (Section~\ref{sec:attribution_method}). 
We select the 2025 models for the main comparison so that papers in
the 2026 corpus postdate their release, reducing the likelihood of
prior exposure to these papers during training. 


All models receive the same ordered literature sets, with papers represented by titles and abstracts. We generate 45,000 ideas across all models and conditions, each containing a title, motivation, core idea, approach, and validation. Additional results for closed-source models and for reasoning-enabled generation are reported in Appendices~\ref{app:closed-models-retrieval} and ~\ref{app:reasoning_ideation}, respectively.

\subsection{Evaluation Metrics}
\label{sec:evaluation}

We evaluate quality, novelty, and diversity as complementary dimensions of
creative generation.
Individually novel ideas can
still concentrate on closely related questions or methods. We therefore
assess both individual ideas and the breadth of the generated collection. Below, we define the corresponding evaluation measures.

\paragraph{Diversity.}
We measure semantic diversity as the mean pairwise cosine distance
between \texttt{text-embedding-3-large} embeddings
\citep{openai2024embeddings} of each idea.
Within-set diversity averages this measure over five-idea generation
groups, each generated by the same model from the same literature
set $\mathcal{D}$.
Overall diversity applies the same measure to all 50 ideas per
model--topic--condition combination.
We use these measures to compare variation within generation groups
with variation across ideas pooled from different groups.

\paragraph{Quality.}
We adopt the validity, feasibility, significance, and clarity dimensions
from InnoEval~\citep{qiao2026innoeval}.
\texttt{GPT-5.6-Sol} scores each dimension on a 1--10 scale, and their
arithmetic mean gives the quality score.
We evaluate five ideas per model--topic--condition combination.

\paragraph{Novelty.}
\texttt{GPT-5.6-Luna} scores novelty using the five-point RINoBench
rubric~\citep{schopf2026rinobench}.
Local novelty compares an idea with its supplied literature $\mathcal{D}$
and is evaluated for all literature-conditioned ideas.
Search-assisted global novelty compares an idea with additional related
papers retrieved through web search. It is evaluated on one idea per
model--topic--condition combination, including \cnolit.

Full agreement analyses and judge-selection results are reported in Appendix~\ref{app:agreement_across_judges}.
Evaluation prompts are provided in Appendix~\ref{app:judge_prompts}; judge-agreement results and analyzes are reported in Appendix C.

\paragraph{Human evaluation.}
Three PhD students with an NLP background evaluate 20 ideas
sampled from the NLP topic using the same rubric as the
LLM judge.
For each idea, both the annotators and the judge receive the
eight retrieved input papers and the five most recent papers retrieved by GPT-5.6-Sol with web-search tool.
The judge's Spearman correlations with mean human ratings are
0.307 for quality, 0.680 for local novelty,
and 0.504 for global novelty.
Appendix~\ref{app:human_evaluation} provides the evaluation
details and agreement between human annotators.

\begin{figure*}[ht]
    \centering
    \includegraphics[width=0.95\textwidth]{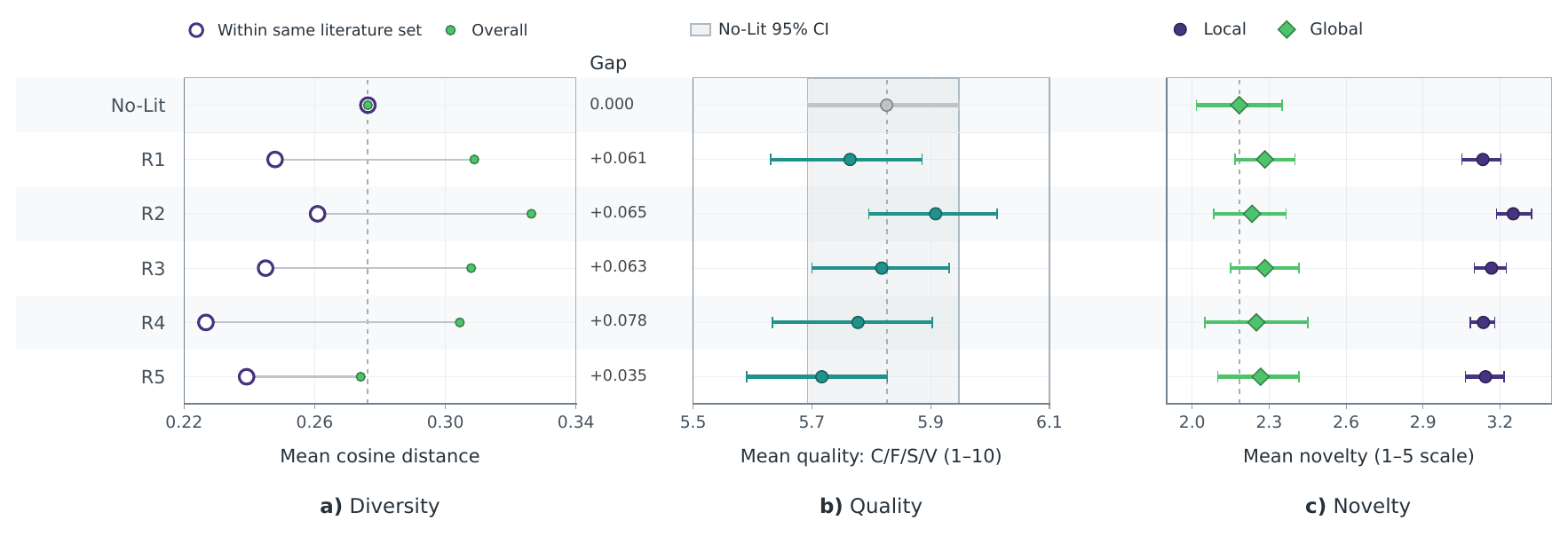} 
    \caption{
    \textbf{Retrieval changes the range of scientific exploration more than
    quality and novelty.}
    (a) Mean cosine distance among ideas generated from the same literature
    set (open circles) and across all sets under each retrieval strategy
    (green). ``Gap'' denotes overall minus within-set diversity; for
    \cnolit the two coincide.
    (b) Mean quality with 95\% confidence intervals; the shaded region marks
    the 95\% confidence interval of \cnolit.
    (c) Mean local and global novelty with 95\% confidence intervals.
    }
    \label{fig:main-results}
\end{figure*}

\begin{figure*}[t]
    \centering
    \includegraphics[width=0.9\linewidth]{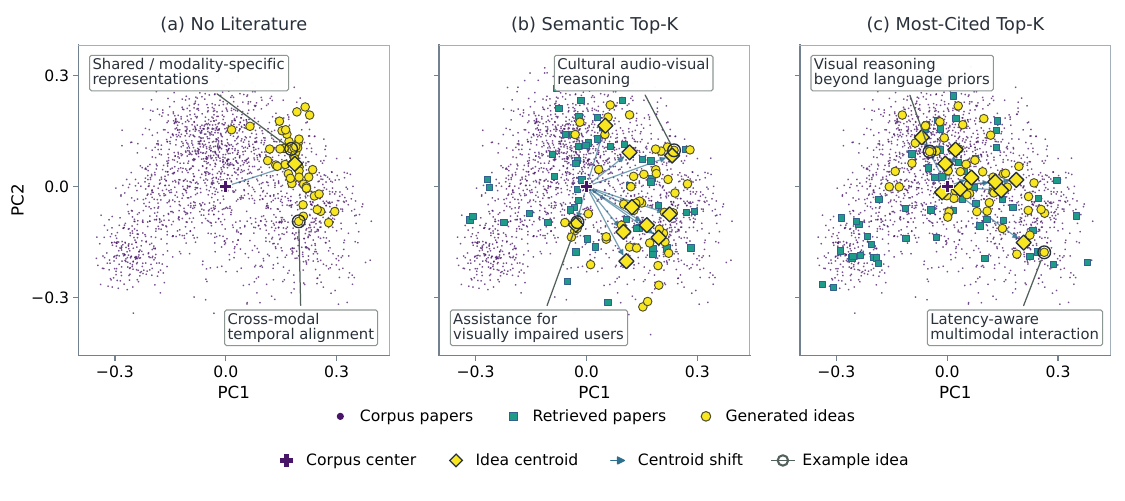}
\caption{\textbf{Ideas spread more widely with retrieval in the PCA visualization.} Each panel shows 50 Qwen3 ideas and 2,000 corpus papers for the topic \emph{Vision, Language, Audio, and Multimodal Applications}.}
    \label{fig:figure-1-multimodal-semantics}
\end{figure*}

\begin{figure*}[t]
  \centering
  \includegraphics[width=0.8\textwidth]{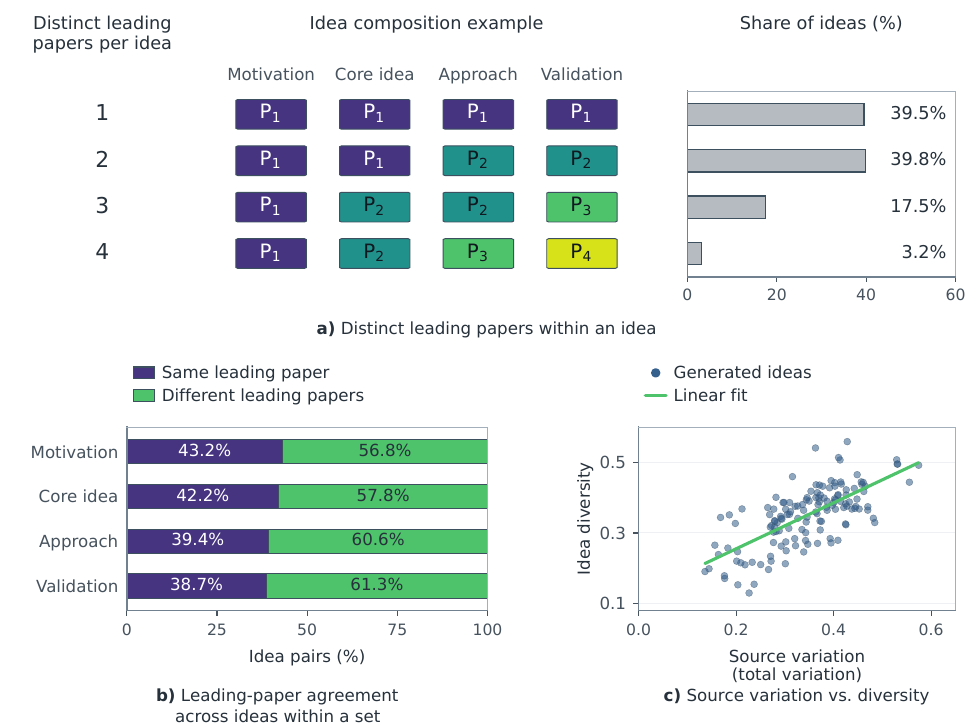}
  \caption{
    \textbf{Leading-source reuse and idea diversity.}
    (a) In $79.3\%$ of ideas, at most two papers lead the four components.
    (b) Within a fixed literature set, $38.7\%$--$43.2\%$ of idea pairs share
    the same leading paper for a given component, over three times the
    $12.5\%$ baseline under independent, uniform source selection.
    (c) Across groups of 20 ideas, greater source variation is associated
    with higher idea diversity (Pearson $r=0.70$).
  }
  \label{fig:source_dependence}
\end{figure*}

\section{Findings}
\label{subsec:rq1}
\subsection{How does retrieved literature shape scientific exploration?}
\label{sec:retrieval-results}

We first compare the five retrieval strategies across the four 2025
models and 15 research topics. Figure~\ref{fig:main-results} reports
diversity within a fixed literature set and across all literature sets,
together with quality and novelty.

\paragraph{Retrieval narrows diversity among ideas generated from the same papers but can broaden it across different sets of papers.}
All five retrieval strategies reduce diversity among ideas generated
from the same papers relative to \cnolit, with decreases of
$5.6\%$--$17.9\%$ (Figure~\ref{fig:main-results}a).
When ideas generated from different sets of papers are considered
together, \crone--\crfour increase \coverall diversity by
$10.2\%$--$18.1\%$ relative to \cnolit, while \crfive remains
close to the baseline ($-0.8\%$).
The gap between overall and within-set diversity is largest
for \crfour ($+0.078$) and smallest for \crfive ($+0.035$).
\crfive also has the lowest mean distance between the centroids
of its input paper sets (0.046, compared with 0.066--0.115
for \crone--\crfour).
After accounting for topic and retrieval strategy, this distance
is positively associated with overall idea diversity ($r=0.503$)
and the gap between overall and within-set diversity ($r=0.356$).
These associations suggest that selecting semantically different
sets of papers can help broaden exploration.

\paragraph{Differences in exploration are larger than differences in
average idea quality and novelty.}
Retrieval produces comparatively modest changes in proposal-level scores
(Figure~\ref{fig:main-results}b,c). \crtwo obtains the highest mean quality
($5.9$) and local novelty ($3.3$), but no retrieval strategy consistently
dominates across quality and novelty. Mean global-novelty scores vary little
across retrieval strategies. Strategies with substantially different diversity can therefore have similar average proposal-level scores.

\paragraph{Retrieval effects vary across research topics.}
The aggregate results vary substantially across topics
(Appendix~\ref{app:topic-retrieval-preferences}). The strategy producing the highest
quality or novelty differs across topics, indicating that no single
literature retrieval strategy is uniformly preferable. The difference between within-context and across-context diversity is more
consistent across topics than the ranking of individual retrieval strategies.
Figure~\ref{fig:figure-1-multimodal-semantics} illustrates this difference
for \emph{Vision, Language, Audio, and Multimodal Applications}. Ideas
generated without literature occupy a comparatively concentrated region,
whereas different retrieved literature sets shift generations toward
different regions of the projected space.

\paragraph{The within- versus across-context difference extends across idea
components.}
The same pattern is observed when diversity is computed separately for the
title, motivation, core idea, approach, and validation plan
(Appendix~\ref{app:rq1-component-diversity}). For every retrieval strategy,
overall diversity exceeds diversity among ideas generated from the same
literature set. The difference is largest under \crfour, where, for example,
motivation diversity increases from $0.274$ within a literature set to
$0.369$ across sets. The same pattern therefore holds across all proposal components.

\subsection{How do retrieval effects change across model updates?}
\label{sec:temporal-results}

The preceding analysis holds the generation model fixed when comparing
literature contexts. We next examine whether the effect of retrieval itself
changes as the underlying model changes. We compare matched 2025 and 2026
checkpoints from four model families while holding topics and ordered
literature sets fixed.

\paragraph{Retrieval provides less additional diversity in newer models.}
Diversity under \cnolit increases by $0.027$ from 2025 to 2026,
whereas average diversity under retrieval changes little
(Figure~\ref{fig:model-updates}a). Consequently, the average difference
between retrieval and \cnolit changes from $+0.015$ in 2025 to
$-0.015$ in 2026. This $0.030$ reduction in the retrieval effect is driven
primarily by greater diversity without retrieved literature, rather than by
a corresponding decline in retrieval-conditioned diversity.

\paragraph{The relative effect on quality changes in the opposite direction.}
Mean quality decreases by $0.343$ under \cnolit, compared with an
average decrease of approximately $0.094$ under retrieval. The quality
difference between retrieval and \cnolit consequently increases by
approximately $0.249$. Across these model updates, retrieved literature
therefore provides less additional variation but a larger relative advantage
in quality.

\begin{figure*}[ht]
  \centering
  \includegraphics[width=0.8\textwidth]{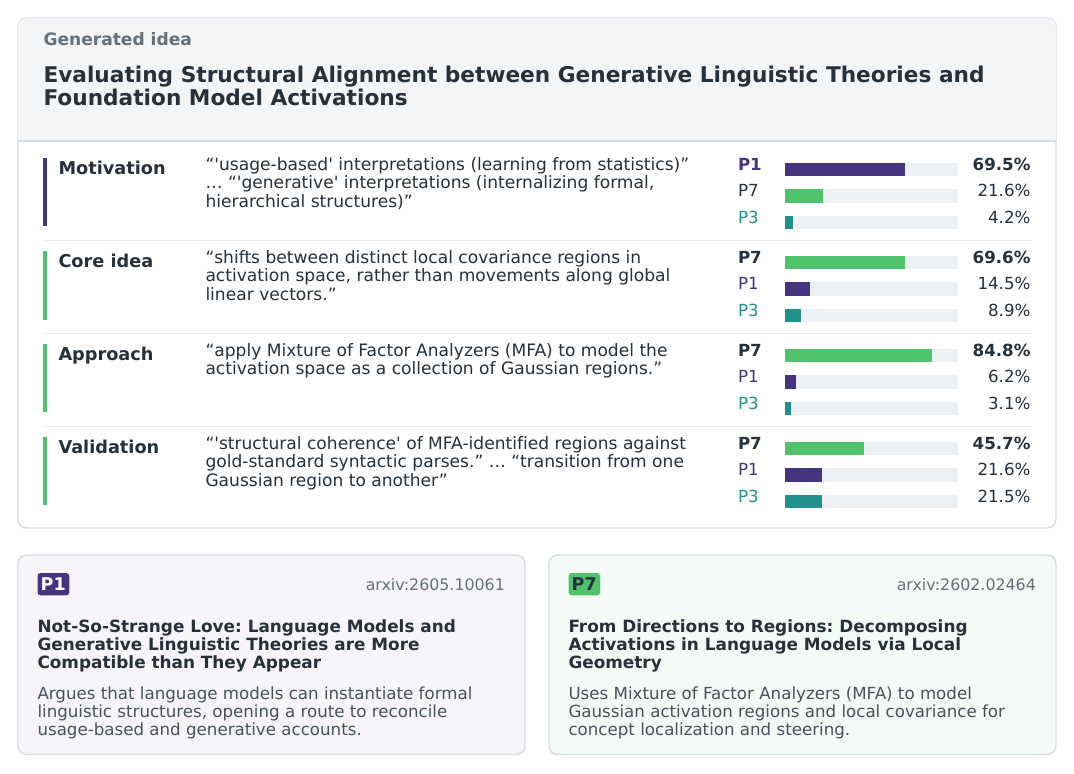}
  \caption{\textbf{Different leading sources within one generated idea.}
  P1 leads the motivation; P7 leads the core idea, approach, and validation.
  Bars show the top three shares of mean token-level leave-one-out JSD, normalized over all eight input papers per component.
  Colored accents identify each component's leading paper.}
  \label{fig:rq2-nlp-component-case}
\end{figure*}

\paragraph{Changes are model-family specific.}
The changes differ across model families
(Figure~\ref{fig:model-updates}b). Gemma gains in quality while decreasing
in local novelty, whereas GLM and Nemotron gain in local novelty while
decreasing in quality; Nemotron also becomes less diverse. Qwen shows the
clearest increase in exploration, with gains in both diversity and local
novelty. Search-assisted global novelty does not change significantly for
any family. Improvements in one dimension of scientific ideation therefore
do not imply corresponding improvements in the others.

\section{How do generated ideas depend on retrieved papers?}
\label{subsec:rq2}
\label{subsec:attributio}

The preceding analyses show that changing the literature broadens the idea
space and that the magnitude of this effect varies across models. We now turn
to the fixed-context setting to examine how generated ideas depend on the
individual papers within a retrieved set.

We randomly select five of the 15 topics and increase generation from five to 20 ideas per fixed model--literature input.
Figure~\ref{fig:source_dependence} summarizes source reuse across components and ideas, along with the association between source variation and idea diversity. Figure~\ref{fig:rq2-nlp-component-case} illustrates how the leading paper differs across components within one idea. Appendix~\ref{app:attribution_robustness} compares attribution
with human judgments and model-reported sources.
Additional results across model updates and a qualitative study are presented in
Appendices~\ref{app:temporal} and~\ref{app:source_case_study}, respectively.

\paragraph{A few papers lead across components and recur across ideas.}
In $79.3\%$ of ideas, one or two papers lead across motivation, core idea, approach, and validation; a single paper leads all four in $39.5\%$ (Figure~\ref{fig:source_dependence}a). Across ideas sharing the same literature, $38.7\%$--$43.2\%$ of pairs share a leading paper for a given component, over three times the $12.5\%$ expected under uniform selection (Figure~\ref{fig:source_dependence}b). Source dependence is therefore concentrated on a few papers,
with the same leading sources recurring across ideas generated
from the same papers.

\paragraph{More varied source dependence accompanies greater idea diversity.}
Groups with more varied core-idea source dependence contain more diverse ideas (Pearson $r=0.70$, $95\%$ CI $[0.56,0.81]$; Figure~\ref{fig:source_dependence}c). Figure~\ref{fig:rq2-nlp-component-case} shows how source dependence can shift across components of the same idea. P1~\citep{mccoy2026not} leads the motivation ($69.5\%$), while P7~\citep{shafran2026directions} leads the core idea ($69.6\%$) and approach ($84.8\%$). Validation is more distributed across P7 ($45.7\%$), P1 ($21.6\%$), and P3 ($21.5\%$). 
This case highlights the distinct roles retrieved papers can play within a single idea, revealing component-specific patterns of literature influence.

\paragraph{Presentation position strongly influences attribution.}
Even after shuffling the same eight papers, the first displayed paper received $21.7\%$--$34.5\%$ of attribution across four models, compared with $12.5\%$--$16.4\%$ for the next-highest position in each model. 
Doubling title--abstract length was also associated with a $4.62$-percentage-point increase in attribution share, while paper type and topic relevance showed no clear independent associations.

\paragraph{Implications for literature-grounded ideation.}
Our results suggest several practical considerations. Generating more ideas
from the same literature does not recover the breadth obtained by varying the
literature across generations (\S\ref{subsec:rq1}). Retrieval strategies should
therefore be evaluated over sets of ideas, since similar average quality or
novelty can coincide with substantial differences in diversity
(\S\ref{subsec:rq1}). Recurring dependence on the same papers is also associated
with more concentrated idea sets (\S\ref{subsec:rq2}), although whether
deliberately varying source dependence can broaden exploration remain to be
tested. Finally, retrieval effects change across model updates
(\S\ref{sec:temporal-results}): newer models may exhibit greater baseline diversity without
retrieved literature while retaining a relative quality advantage from
retrieval. Retrieval strategies should therefore be re-evaluated as the
underlying generation model changes, rather than assumed to transfer
unchanged across model generations.

\section{Related Work}
\label{sec:related_work}

\paragraph{Literature-grounded scientific ideation.}
Scientific ideation systems retrieve literature to guide idea generation and refinement \citep{eger2026thaisci}. SciMON and ResearchAgent combine retrieval with iterative refinement \citep{wang2024scimon,baek2025researchagent}, while SciPIP combines semantic, entity, and co-citation retrieval \citep{wang2024scipip}. Chain of Ideas and Graph2Idea organize literature through developmental chains and graph relations \citep{li2024chain,li2026graph2idea}. FlowPIE couples feedback-guided patent literature exploration with idea evolution \citep{wang2026flowpie}; EvoGens uses retrieval-guided mutation and semantic-aware crossover \citep{li2026evogens}. We compare retrieval strategies under a shared generation protocol across models and successive releases.

\paragraph{Scientific exploration and evaluation.}
NoveltyBench and Artificial Hivemind document limited diversity in open-ended generation \citep{zhang2025noveltybench,jiang2025hivemind}. In scientific ideation, \citet{tang2026narrow} find that generated ideas remain concentrated around seed literature. Heuresis finds that search strategies change quality, diversity, and novelty without expanding the observed quality--novelty frontier \citep{antoniades2026heuresis}, while IDEAgent counts distinct ideas meeting quality thresholds \citep{gumma2026ideagent}. We distinguish diversity within a fixed paper set from diversity across retrieved sets and evaluate idea quality and novelty. Novelty judgments use supplied papers and, following evidence-grounded approaches such as InnoEval \citep{qiao2026innoeval}, additional literature retrieved through web search.

\paragraph{Attributing generation to retrieved literature.}
ALCE evaluates whether cited evidence supports generated claims \citep{gao2023alce}. ContextCite and ARC-JSD estimate source influence through context ablations, measuring changes in response likelihood and output distributions, respectively \citep{cohenwang2024contextcite,li2025arcjsd}. However, how models rely on retrieved literature during scientific ideation remains unclear. We investigate these relationships by tracing paper-level influence within and across generated ideas.

\section{Conclusion}
\label{sec:conclusion}
We study how retrieved literature shapes scientific ideation under a controlled generation protocol. Varying the retrieved literature generally broadens the range of research directions explored, while ideas generated from a fixed literature set remain more concentrated. This effect is stronger for diversity than for average quality or novelty. Paper-level attribution reveals concentrated reliance on a few retrieved papers, recurring leading sources across generations, and substantial variation in attribution with paper presentation position. These patterns also vary across model updates. The literature supplied to a model affects both individual ideas and the distribution of research directions produced across generations. Future work should evaluate literature-grounded ideation at the level of idea sets, considering the range of directions explored alongside quality and novelty.
\section*{Limitations}

Our diversity measures rely on distances between text embeddings, which do not necessarily correspond to substantive differences in research questions, underlying mechanisms, or methods. Ideas with different wording or framing may still share the same scientific approach. Search-assisted global novelty depends on the coverage of retrieved prior work: relevant studies may be missed, leading to overestimated novelty. Quality and novelty judgments also depend on the LLM judges, and agreement across providers alone does not establish agreement with domain experts. Global novelty is evaluated only on a subset of ideas, limiting the precision of comparisons for this metric, particularly when score differences are small.

The experiments cover 15 ICLR-derived topics, a fixed literature corpus, and selected open-weight model families, with each literature set providing the titles and abstracts of eight papers. Results may differ for other disciplines, model scales, full-text inputs, or iterative ideation systems. Papers published after a checkpoint's release may still contain concepts encountered during training. Within-family comparisons also combine changes in training and model design, so release year alone cannot explain generational differences.

Our attribution analysis removes papers one at a time and compares model predictions for the same idea text, with preceding tokens held fixed. The resulting attribution scores reflect changes in the model's next-token distributions, rather than establishing the scientific origins of ideas or fully capturing interactions among retrieved papers. Although additional checks support the correspondence between attribution scores and model-reported sources (Appendix~\ref{app:attribution_robustness}), these reports do not provide independent ground truth for source influence. The association between attribution patterns and idea diversity is correlational. We evaluate ideas without implementing them, so their practical scientific value remains untested.

\section*{Ethics statement}

Our study uses publicly available scientific literature and model-generated research ideas. Generated ideas may reproduce prior work, contain scientific errors, or suggest harmful applications, and require expert assessment before implementation. Literature selection may also overrepresent highly cited work and established research communities. Responsible use should preserve source provenance, credit prior contributions, and assess the potential consequences of pursuing an idea. 

\section*{Reproducibility statement}

Section~\ref{sec:ideation_protocol} defines the generation protocol. Section~\ref{sec:experimental_setup} specifies the corpus construction, model comparisons, and evaluation measures. Section~\ref{sec:attribution_method} defines the attribution method. Appendix~\ref{app:exp_protocol} details checkpoint identifiers, generation and judge settings, evaluation subsets and sampling procedures, and statistical protocols. Reproduction requires preserving the ordered literature contexts shared across checkpoints and the idea subsets used for each evaluation. Details of artifact licensing and our release plan are provided in Appendix~\ref{app:licenses}.

\section*{AI Use Statement}

In this study, we used LLMs for research idea generation, quality and novelty evaluation, and attribution analysis, and text embedding models to measure idea diversity. These uses are part of the experimental protocol described in the paper.

We also used generative AI tools to assist with manuscript editing, coding, and figure design. We did not use these tools to formulate this study's research questions or hypotheses, interpret its experimental results, or develop its scientific claims. All AI-assisted edits and figures were reviewed and revised by the authors. We take responsibility for the final content of this work, including any text or artifacts produced with the aid of generative AI.

\bibliography{custom}

\appendix

\appendix
\section{Experimental Protocol}
\label{app:exp_protocol}

\subsection{Models and Comparisons}
\label{app:models}

Table~\ref{tab:model_roster} lists the ten open-weight models
and three proprietary models used in the study.
The main comparison of retrieval strategies uses the four
2025 open-weight models. We compare each with its 2026
counterpart from the same family, with additional
2024--2026 comparisons for Qwen and Gemma.
Models receive the same topics, literature sets, paper order,
and generation task for each matched comparison.

\begin{table}[t]
    \centering
    \begin{tabular}{llr}
        \hline
        Group & Model & Parameters \\
        \hline
        2024 & Qwen2.5 & 32B \\
             & Gemma 2 & 27B \\
        \hline
        2025 & Qwen3-2507 & 30B \\
             & Gemma 3 & 27B \\
             & Nemotron 3 Nano & 30B \\
             & GLM-4 & 32B \\
        \hline
        2026 & Qwen3.6 & 35B \\
             & Gemma 4 & 31B \\
             & Nemotron Cascade 2 & 30B \\
             & GLM-4.7-Flash & 30B \\
        \hline
        Proprietary & GPT-5.6 Luna & -- \\
                    & Claude Haiku 4.5 & -- \\
                    & Gemini 3.5 Flash-Lite & -- \\
        \hline
    \end{tabular}
    \caption{Models used in the study. Open-weight models
    are grouped by release year.}
    \label{tab:model_roster}
\end{table}

\subsection{Topics, Corpus, and Literature Sets}
\label{app:corpus_contexts}

We construct 15 research topics from the 21-area ICLR 2026
taxonomy by merging related areas and excluding
``Other Machine Learning.''
Table~\ref{tab:research_topics} lists the topics and their
constituent areas.
Eligible papers were first submitted to arXiv between
January 1 and July 23, 2026.
For each topic, we select the 2,000 most semantically relevant
papers. Deduplicating these selections yields a shared corpus
of 24,836 papers.
Each literature set contains eight papers, represented by
their titles and abstracts.

\begin{table}[t]
    \centering
    \small
    \renewcommand{\arraystretch}{1.15}
    \begin{tabular}{|p{0.45\linewidth}|p{0.45\linewidth}|}
        \hline
        \textbf{Research topic} & \textbf{Constituent areas} \\
        \hline
        Representation, Transfer, Meta, and Continual Learning
        & Representation learning; transfer, meta, and lifelong learning \\
        \hline
        Foundation and Generative Models
        & Foundation/frontier models; generative models \\
        \hline
        Reinforcement Learning and Decision Making
        & Reinforcement learning \\
        \hline
        Probabilistic and Causal Learning
        & Probabilistic methods; causal reasoning \\
        \hline
        Optimization and Learning Theory
        & Optimization; learning theory \\
        \hline
        Graph and Geometric Learning
        & Graphs, geometries, and topologies \\
        \hline
        Time-Series and Dynamical Learning
        & Time series and dynamical systems \\
        \hline
        Alignment, Safety, Fairness, Privacy, and Societal Considerations
        & Alignment, fairness, safety, privacy, and society \\
        \hline
        Interpretability and Explainable AI
        & Interpretability and explainable AI \\
        \hline
        Datasets and Benchmarks
        & Datasets and benchmarks \\
        \hline
        ML Infrastructure and Systems
        & Infrastructure, software, hardware, and systems \\
        \hline
        Neurosymbolic and Hybrid AI
        & Neurosymbolic and hybrid AI \\
        \hline
        Vision, Language, Audio, and Multimodal Applications
        & Multimodal applications \\
        \hline
        Robotics, Autonomy, and Planning
        & Applications to robotics, autonomy, and planning \\
        \hline
        ML for Scientific and Cognitive Domains
        & Applications to neuroscience and cognitive science;
        applications to physical sciences \\
        \hline
    \end{tabular}
    \caption{The 15 research topics and their constituent
    ICLR 2026 areas. Area names are abbreviated.
    }
    \label{tab:research_topics}
\end{table}

We use the five retrieval strategies defined in
Section~\ref{sec:ideation_protocol}: Semantic Top-K,
Most-Cited Top-K, Semantic + Entity + Co-citation,
Core-Paper Reference Expansion, and Chain of Ideas.
For Semantic Top-K and Most-Cited Top-K, each literature set
is sampled from the strategy's top 100 candidates.
The other strategies select papers through their respective
retrieval procedures.
The \cnolit{} condition supplies only the research topic.

\subsubsection{Retrieval Implementation}
\label{app:retrieval-implementation}

We implement the five strategies in Section~\ref{sec:retrieval} on the shared corpus $P$.
Let $P_q$ denote the 2,000 papers retrieved for topic $q$.
We encode the topic name and each paper's title and abstract with
\texttt{text-embedding-3-large} and compute cosine similarity between
their 3,072-dimensional embeddings.
Citation counts and reference links come from a fixed Semantic Scholar
snapshot; citation-based expansion uses links within $P$.
Each strategy produces ten eight-paper inputs per topic, whose contents
and order are shared across models.

\paragraph{R1: Semantic Top-K.}
We retain the 100 papers in $P_q$ with the highest topic similarity
and sample eight uniformly without replacement.
We repeat sampling to obtain ten distinct sets, redrawing duplicate
sets while allowing individual papers to recur across sets.
Selected papers retain their semantic ranking in the input.

\paragraph{R2: Most-Cited Top-K.}
We rank papers in $P_q$ by citation count, using semantic rank and
then arXiv ID to break ties.
After deduplicating by Semantic Scholar work identity, we retain
the first 100 papers and apply the sampling procedure used for R1.
Selected papers retain their citation ranking in the input.

\paragraph{R3: Semantic + Entity + Co-citation.}
We combine three candidate sets.
The semantic set contains the five papers most similar to the topic.
For entity retrieval, GPT-5.4-mini extracts up to five scientific
entities per topic and twelve per paper, covering tasks, methods,
datasets, and evaluation measures.
Orthographic variants, singular/plural forms, and explicitly defined
abbreviations are normalized.
Entities from the topic and the five semantic candidates are expanded
with up to ten additional entities that each co-occur with an initial
entity in at least seven corpus papers, preferring those with lower
document frequency.
Matching papers form the entity set.
We cap this set at 500 papers by successively halving the expansion
count.
If the initial entities alone exceed this cap, we retain the 500 most
topic-relevant papers matching the initial entities and the original
expansion.
For each semantic or entity candidate, the co-citation set adds the
paper with the most shared citing papers in $P$, requiring at least
two shared citing papers.

We merge the candidates and apply single-link clustering at cosine
similarity $0.8$.
Clusters are ranked by their most topic-relevant member.
For each input, we uniformly sample eight of the first 100 clusters
without replacement and one paper uniformly from each selected cluster.
We redraw duplicate sets and present papers in cluster-rank order.

\paragraph{R4: Core-Paper Reference Expansion.}
We rank potential cores by topic similarity, considering $P_q$ before
the remaining corpus and prioritizing papers with references in $P$.
We select ten distinct works mapped to Semantic Scholar, one core
per input.
Each core is paired with up to seven references, ranked by topic
similarity.
Remaining positions are filled from the semantic ranking of $P_q$,
followed by $P$ if needed, excluding already selected works.
Inputs present the core first, then its references, followed by the
additional semantic selections.

\paragraph{R5: Chain of Ideas.}
We sample two distinct anchors without replacement from mapped works
among the 50 most topic-relevant papers in $P_q$, with probabilities
proportional to nonnegative topic similarity.
Each anchor starts a chain of up to four papers.
At each step, we compare the best-scoring reference at the backward
end with the best-scoring citing paper at the forward end and extend
toward the higher-scoring candidate, excluding works already in that
chain and the other anchor.
Expansion ends at four papers or when no eligible neighbor remains.
We merge the two chains, deduplicate by work identity, and fill
remaining positions with the most topic-relevant unselected mapped
works in $P$.
Inputs are ordered by publication year and then arXiv ID.

\subsection{Generation and Sampling}
\label{app:generation_settings}

The ten open-weight models generate ideas at temperature
$T=1$. Each idea is generated in a separate call and contains
five fields: title, motivation, core idea, approach, and
validation plan.

For each model, topic, and retrieval strategy, we use ten
literature sets and generate five ideas from each set,
giving 50 ideas per condition.
The \cnolit{} condition generates 50 ideas per model and
topic using the same topic-only prompt.
Across 15 topics and six conditions, this gives 4,500 ideas
per model and 45,000 across the ten open-weight models.
Of these, 37,500 use retrieved literature and 7,500 belong
to \cnolit{}.

For the expanded attribution analysis, each group contains
20 ideas generated by the same model from the same
literature set.

\subsection{Evaluation Coverage}
\label{app:evaluation_coverage}

Table~\ref{tab:evaluation_coverage} summarizes evaluation
coverage for the open-weight models.
Quality is evaluated along the four dimensions defined in
Section~\ref{sec:experimental_setup}: clarity, feasibility,
significance, and validity.
Each dimension receives a score from 1 to 10, and their
mean gives the overall quality score.

Local and search-assisted global novelty use the five-point
rubric described in the same section.
Local novelty is assessed relative to the supplied papers.
Global novelty is assessed against literature found through
web search and also covers \cnolit{} ideas.
For judge agreement, participating judges evaluate the
same ideas.

\begin{table}[t]
    \centering
    \begin{tabular}{p{0.32\linewidth}p{0.59\linewidth}}
        \hline
        Analysis & Coverage \\
        \hline
        Generation and embeddings
        & 45,000 ideas across ten open-weight models \\
        Quality
        & Five ideas per model, topic, and condition;
        4,500 ideas in total \\
        Local novelty
        & All 37,500 ideas generated with retrieved literature \\
        Search-assisted global novelty
        & One idea per model, topic, and condition;
        900 ideas in total \\
        Attribution
        & Ten open-weight models; expanded groups contain
        20 ideas from the same model and literature set \\
        \hline
    \end{tabular}
    \caption{Evaluation coverage for the open-weight models.
    A condition is one of the five retrieval strategies
    or \cnolit{}.}
    \label{tab:evaluation_coverage}
\end{table}

\subsection{Diversity Measures}
\label{app:diversity_details}

We concatenate each idea and embed
the text with \texttt{text-embedding-3-large}.
For $N$ unit-normalized embeddings
$\{\mathbf{z}_i\}_{i=1}^{N}$, diversity is the mean pairwise
cosine distance:
\begin{equation}
    D =
    \frac{2}{N(N-1)}
    \sum_{i<j}
    \left(1-\mathbf{z}_i^\top\mathbf{z}_j\right).
    \label{eq:pairwise_diversity}
\end{equation}

For each model, topic, and condition, \emph{overall diversity}
uses all 50 ideas.
\emph{Within-set diversity} is computed among ideas generated
from each literature set and then averaged across the ten sets.
For \cnolit{}, groups share the same topic-only prompt.
We also measure the mean cosine distance between ideas
generated from different literature sets.

Additional measures include the mean distance from each
idea to its nearest neighbor and the Vendi Score.

\subsection{Tracing Literature Influence}
\label{app:attribution_details}

We compare the generating model's predictions for the same
idea text with all eight papers present and with each paper
omitted in turn.
For idea $i$, paper $k$, and component $c$, let $A_{i,k,c}$
denote the token-averaged JSD score defined in
Equation~\ref{eq:component_jsd}.
We normalize these scores to obtain the attribution
distribution for each component:
\begin{equation}
    w_{i,k,c} =
    \frac{A_{i,k,c}}
    {\sum_{\ell=1}^{8}A_{i,\ell,c}},
    \label{eq:component_source_share}
\end{equation}
when the denominator is positive.
The leading paper is the paper with the largest attribution
share.
For rank profiles, we sort the shares separately for each
idea and component before averaging.

To examine variation among ideas generated from the same
literature set, we use groups of 20 ideas from the same model.
Within each group, we count the distinct leading papers for
the core-idea component.
For ideas $i$ and $j$, the total variation distance between
their core-idea attribution distributions is
\begin{equation}
    \operatorname{TV}(w_i,w_j) =
    \frac{1}{2}\sum_{k=1}^{8}
    \left|
    w_{i,k,\mathrm{core}}-w_{j,k,\mathrm{core}}
    \right|.
    \label{eq:source_tv}
\end{equation}

We average these distances across idea pairs within each
group and correlate the resulting variation with the
group's diversity, computed over all 20 ideas using
Equation~\ref{eq:pairwise_diversity}.
We also compute the correlation after subtracting the
corresponding model--topic--strategy mean from both
quantities.

\subsection{Temporal Comparisons and Statistical Analysis}
\label{app:temporal_statistics}

\paragraph{Changes across model releases.}
We compare 2025 and 2026 models within each of the four
matched families, using the same topics and literature sets.
For each retrieval strategy, we average metric differences
across the four families and 15 topics.
Family-level summaries average across topics and the five
retrieval strategies.

For metrics also defined under \cnolit{}, the retrieval
effect is the difference between a retrieval condition and
the corresponding \cnolit{} baseline.
We compare changes in this difference across model releases.
Families and topics receive equal weight in aggregate results.

\paragraph{Directions associated with literature sets.}
For each literature set, we subtract the mean embedding of
the model's 50 \cnolit{} ideas from the mean embedding of
ideas generated from that set.
We compare the cosine similarity of these directions across
model releases, using either the same literature set or
different sets within the same family, topic, and retrieval
strategy.
This analysis covers the four strategies with sufficient
distinct literature sets.
Similarities are computed in the original embedding space;
PCA is used for visualization.

\paragraph{Uncertainty and multiple comparisons.}
We use paired-topic bootstrap resampling to estimate the
confidence interval for the change in the retrieval effect
on diversity. Each sampled topic retains its linked
conditions and model releases.
Temporal comparisons use Benjamini--Hochberg correction,
with significance defined by adjusted $q<0.05$.

Global novelty is compared at the model--topic--condition
level because the evaluated literature sets are not fully
matched across model releases.
For correlations, we report 95\% confidence intervals
before and after subtracting the corresponding
model--topic--strategy means.

\section{Prompts for LLM-as-a-Judge Evaluation}
\label{app:judge_prompts}

\begin{figure*}[p]
\centering
\includegraphics[
  width=\textwidth,
  height=0.86\textheight,
  keepaspectratio
]{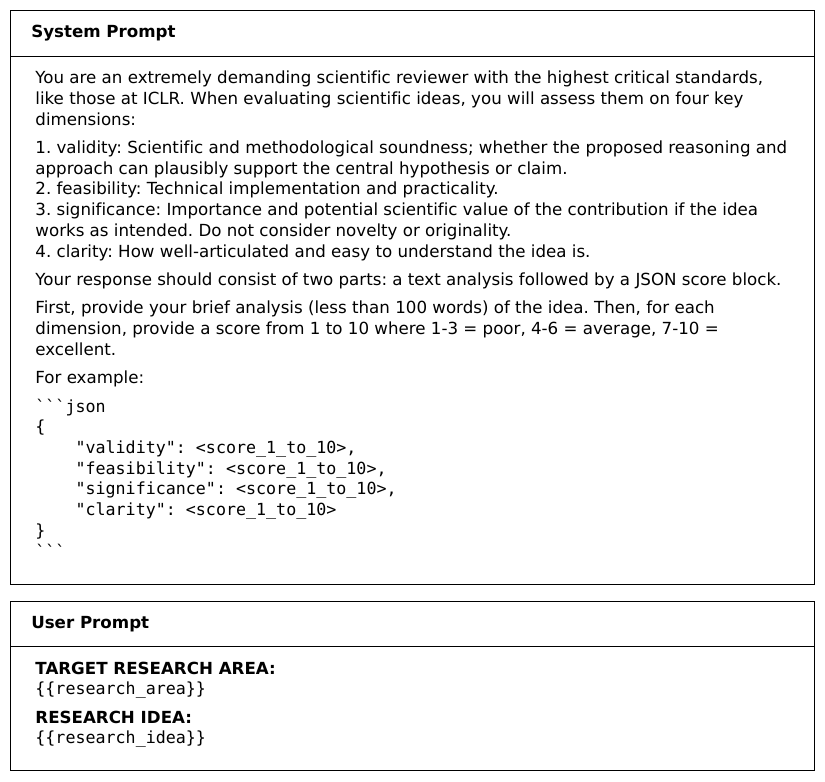}
\caption{Quality evaluation prompt. The judge scores validity,
feasibility, significance, and clarity on a 1--10 scale.}
\label{fig:quality_judge_prompt}
\end{figure*}

\begin{figure*}[p]
\centering
\includegraphics[
  width=\textwidth,
  height=0.86\textheight,
  keepaspectratio
]{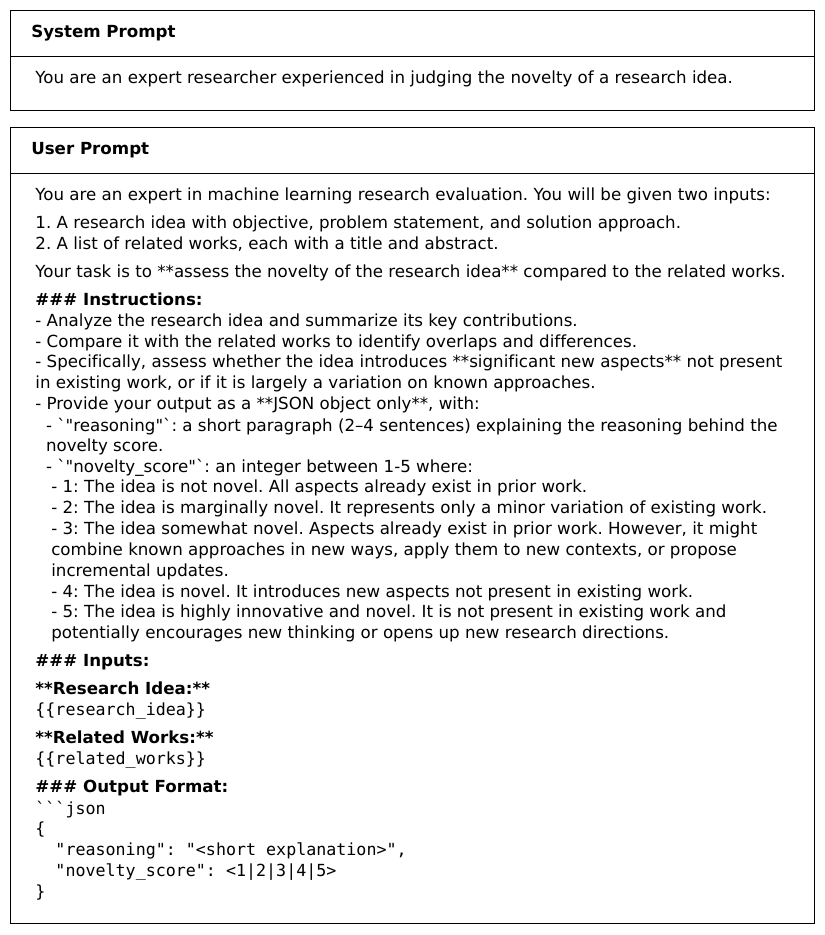}
\caption{Local novelty evaluation prompt. The judge scores the
idea's novelty relative to the supplied papers on a 1--5 scale.}
\label{fig:local_judge_prompt}
\end{figure*}

\begin{figure*}[p]
\centering
\includegraphics[
  page=1,
  width=\textwidth,
  height=0.86\textheight,
  keepaspectratio
]{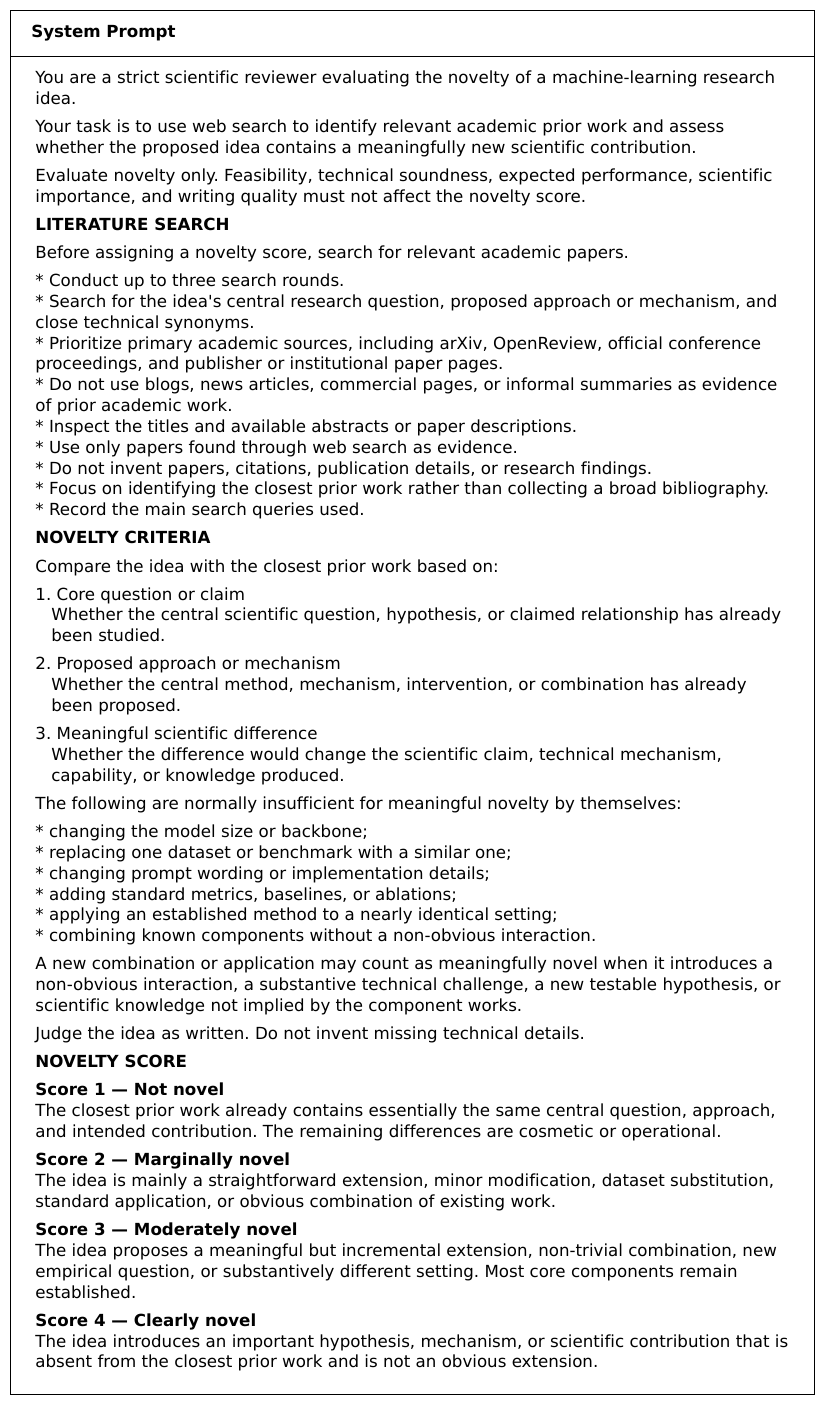}
\caption{Global novelty evaluation prompt (part 1 of 2).
The judge assesses the idea's novelty against related literature
found through web search.}
\label{fig:global_judge_prompt}
\end{figure*}

\begin{figure*}[p]
\centering
\includegraphics[
  page=2,
  width=\textwidth,
  height=0.86\textheight,
  keepaspectratio
]{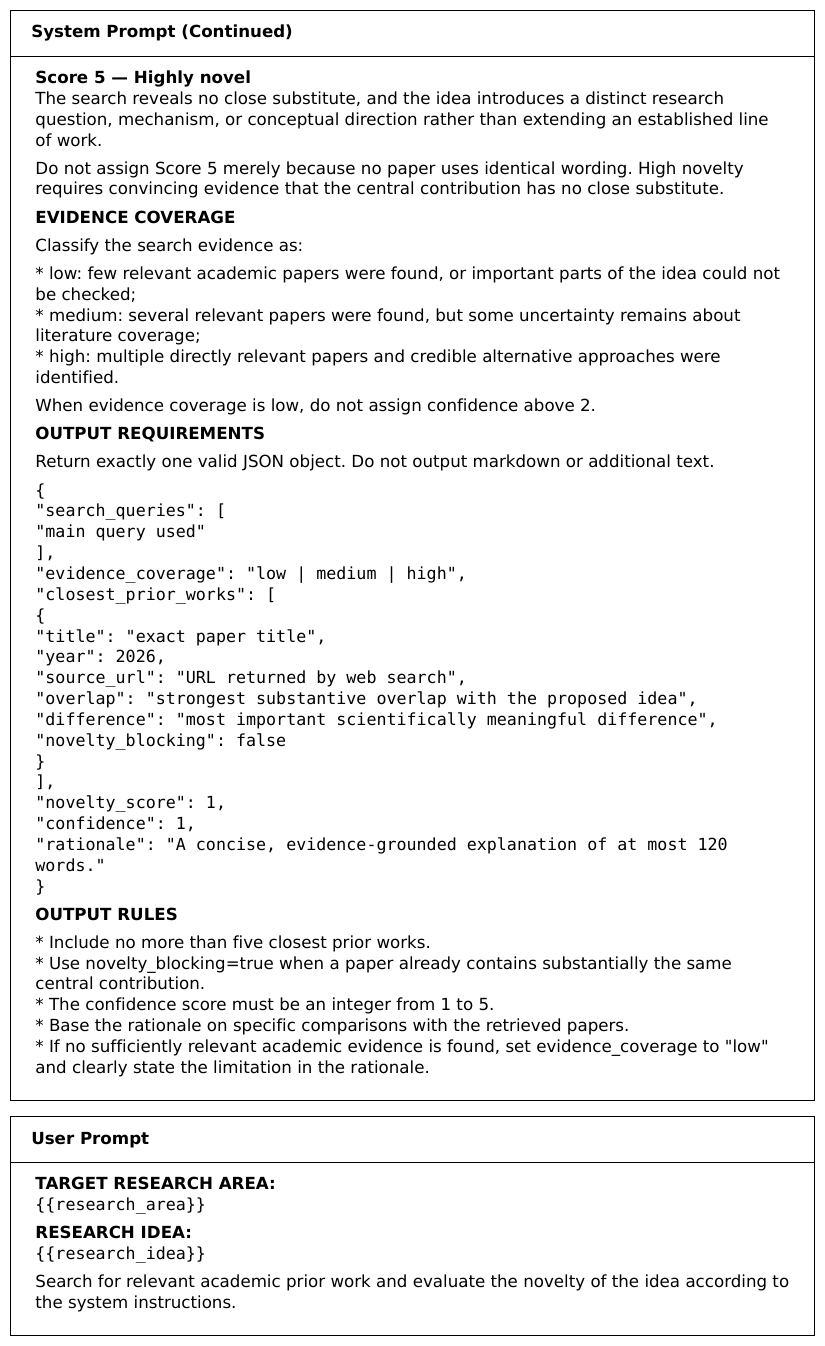}
\caption{Global novelty evaluation prompt (part 2 of 2).}
\label{fig:global_judge_prompt_continued}
\end{figure*}

Figures~\ref{fig:quality_judge_prompt}--\ref{fig:global_judge_prompt_continued}
show the prompts used to evaluate quality, local novelty, and global
novelty. Double-braced fields mark the inputs supplied for each
evaluation. The local novelty prompt follows
RINoBench~\citep{schopf2026rinobench} and is also used to compare
judge scores with the benchmark's reference scores.

\clearpage
\section{Human Evaluation and Agreement Across LLM Judges}
\label{app:agreement_across_judges}

\subsection{Human Annotation Instructions and Interfaces}
\label{app:human_annotation_instructions}

\begin{figure*}[t]
    \centering
    \includegraphics[width=\linewidth]{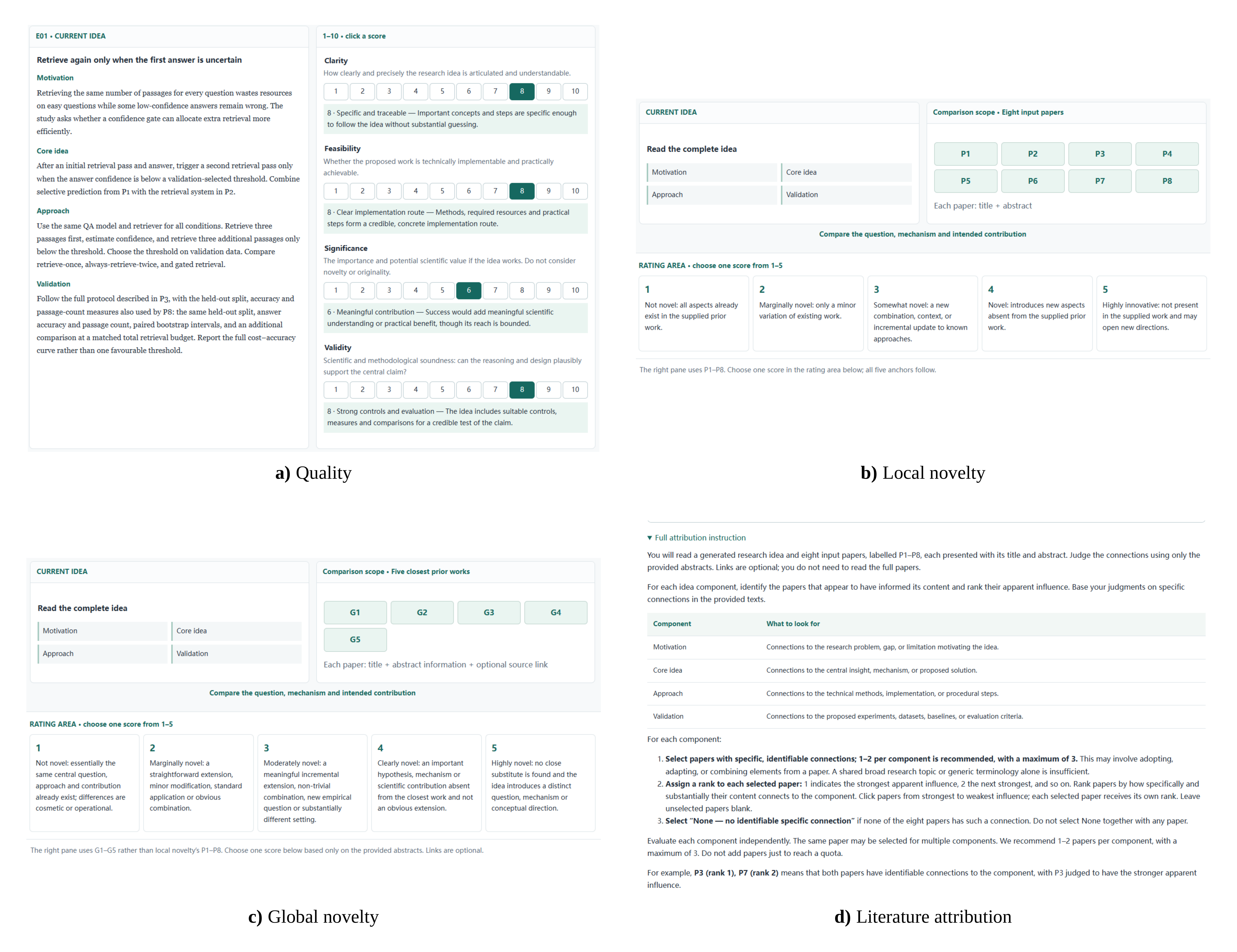}
    \caption{
        Human annotation instructions and interfaces for
        (a) quality,
        (b) local novelty,
        (c) global novelty, and
        (d) literature attribution.
    }
    \label{fig:human_annotation_interfaces}
\end{figure*}

Human annotators used the same rubrics and scoring scales
as the LLM judges for quality, local novelty, and global
novelty. For literature attribution, they identified and
ranked retrieved papers according to their connections to
each idea component.
Figure~\ref{fig:human_annotation_interfaces} shows the
instructions and annotation interfaces.

\subsection{Human Evaluation Results}
\label{app:human_evaluation}

Three PhD students from our research group, all with
backgrounds in NLP, evaluated 20 ideas on NLP.
Participation was voluntary, and annotators received no
additional compensation. They were informed that their
annotations would be used for research and agreed to
this use. For quality, local novelty, and global novelty, we compute
Spearman's rank correlation between the mean human rating
per idea and the score from the primary LLM judge used
for that metric.
Table~\ref{tab:human_evaluation} also reports the mean
pairwise correlation among the three annotators.

\begin{table}[t]
    \centering
    \small
    \begin{tabular}{lcc}
        \toprule
        Metric & Human--Judge & Human--Human \\
        \midrule
        Quality        & 0.307 & 0.501 \\
        Local novelty  & 0.680 & 0.420 \\
        Global novelty & 0.504 & 0.143 \\
        \bottomrule
    \end{tabular}
    \caption{
        Spearman correlations on 20 NLP ideas.
        Human--Judge compares the mean rating from three
        annotators with the primary LLM judge's score.
        Human--Human averages correlations across the
        three annotator pairs.
    }
    \label{tab:human_evaluation}
\end{table}

\subsection{Agreement on Quality}
\label{app:agreement_quality}

We compare GPT-5.6-Sol, Gemini-3.1-Pro-Preview, and
Claude-Opus-5 on a balanced sample of 900 ideas spanning
ten generating models, 15 topics, and six conditions.
All judges evaluate the same ideas using the same rubric.

Table~\ref{tab:quality_judge_agreement} reports quadratically
weighted Cohen's $\kappa$ and two intraclass correlations:
ICC(A,1) measures absolute score agreement, while ICC(C,1)
measures consistency after accounting for fixed differences
in judges' mean scores.

\begin{table}[t]
    \centering
    \small
    \setlength{\tabcolsep}{2pt}
    \begin{tabular}{lrrrrr}
        \toprule
        Score & ICC(A,1) & ICC(C,1)
        & $\kappa_{SG}$ & $\kappa_{SC}$ & $\kappa_{GC}$ \\
        \midrule
        Validity
        & 0.212 & 0.395 & 0.223 & 0.416 & 0.158 \\
        Feasibility
        & 0.574 & 0.616 & 0.546 & 0.545 & 0.616 \\
        Significance
        & 0.103 & 0.479 & 0.401 & 0.068 & 0.083 \\
        Clarity
        & 0.221 & 0.605 & 0.285 & 0.385 & 0.140 \\
        Overall quality
        & 0.196 & 0.549 & 0.338 & 0.218 & 0.139 \\
        \bottomrule
    \end{tabular}
    \caption{
        Agreement on quality ($N=900$).
        S, G, and C denote Sol, Gemini Pro, and Claude Opus.
        All $\kappa$ values use quadratic weights.
        Overall quality scores retain their quarter-point
        spacing.
    }
    \label{tab:quality_judge_agreement}
\end{table}

\subsection{Agreement on Local Novelty}
\label{app:agreement_local_novelty}

\paragraph{Agreement across judges.}
We compare GPT-5.6-Luna, Gemini-3.8-Flash, and
Claude-Haiku-4.5 on 750 ideas from ten generating models,
15 topics, and five retrieval strategies.
Each judge evaluates the same ideas against the same
retrieved papers on a 1--5 scale.
Table~\ref{tab:local_novelty_judge_agreement} reports
ICC(A,1) and pairwise weighted Cohen's $\kappa$.

\begin{table}[t]
    \centering
    \small
    \setlength{\tabcolsep}{6pt}
    \begin{tabular}{ccccc}
        \toprule
        ICC(A,1) & 95\% CI
        & $\kappa_{LG}$ & $\kappa_{LH}$ & $\kappa_{GH}$ \\
        \midrule
        0.654 & [0.607, 0.689] & 0.721 & 0.607 & 0.620 \\
        \bottomrule
    \end{tabular}
    \caption{
        Agreement on local novelty ($N=750$).
        L, G, and H denote Luna, Gemini Flash, and Claude Haiku.
        The ICC confidence interval uses 2,000 bootstrap
        resamples of research topics.
    }
    \label{tab:local_novelty_judge_agreement}
\end{table}

\paragraph{Novelty prediction on RINoBench.}
We evaluate novelty prediction on the 277-item RINoBench
test split~\citep{schopf2026rinobench}, using its ideas,
related papers, and reference scores.
Table~\ref{tab:rinobench_judge_selection} compares archived
benchmark predictions with six GPT-5.6 configurations
that vary in model and reasoning effort.
These include GPT-5.6-Luna with medium reasoning,
which we use for local novelty evaluation.

\begin{table*}[t]
    \centering
    \small
    \setlength{\tabcolsep}{5pt}
    \begin{tabular}{llrrrr}
        \toprule
        Model & Reasoning
        & Accuracy $\uparrow$
        & Macro-F1 $\uparrow$
        & MAE $\downarrow$
        & $\rho$ $\uparrow$ \\
        \midrule
        DeepSeek-R1
        & -- & 0.307 & 0.123 & 0.993 & 0.087 \\
        Llama-3.1-8B
        & -- & 0.289 & 0.146 & 1.004 & -0.014 \\
        Llama-3.3-70B
        & -- & 0.289 & 0.095 & 1.043 & 0.032 \\
        Llama-4-Scout
        & -- & 0.289 & 0.130 & 1.014 & 0.076 \\
        GPT-5
        & -- & 0.289 & 0.172 & 0.928 & 0.140 \\
        GPT-OSS-120B
        & -- & 0.296 & 0.148 & 0.964 & 0.117 \\
        o3
        & -- & 0.303 & 0.162 & 0.931 & 0.113 \\
        \midrule
        GPT-5.6-Luna
        & None & 0.325 & 0.179 & \textbf{0.874} & 0.165 \\
        GPT-5.6-Luna
        & Medium & 0.318 & \textbf{0.234} & 0.924 & 0.166 \\
        GPT-5.6-Terra
        & None & 0.325 & 0.188 & 0.881 & 0.184 \\
        GPT-5.6-Terra
        & Medium & \textbf{0.336} & 0.211 & 0.884 & 0.209 \\
        GPT-5.6-Sol
        & None & 0.321 & 0.187 & \textbf{0.874} & \textbf{0.213} \\
        GPT-5.6-Sol
        & Medium & 0.318 & 0.173 & 0.888 & 0.210 \\
        \bottomrule
    \end{tabular}
    \caption{
        Novelty prediction on RINoBench ($N=277$).
        All models are evaluated against the same reference
        scores. The upper block uses archived benchmark
        predictions; the lower block evaluates GPT-5.6 models
        with the benchmark prompt.
        MAE denotes mean absolute error, and $\rho$ denotes
        Spearman's rank correlation.
        Bold values indicate the best result in each column.
    }
    \label{tab:rinobench_judge_selection}
\end{table*}

\subsection{API Costs}
\label{app:judge_api_cost}

The total estimated API cost of LLM-based evaluation,
including development and validation runs, is \$857.93
(Table~\ref{tab:judge_api_cost}).
Costs combine recorded API charges with estimates based
on usage and published API prices.\footnote{
\url{https://developers.openai.com/api/docs/pricing},
accessed September 24, 2026.}
Reused judgments are counted once.

\begin{table}[t]
    \centering
    \small
    \begin{tabular}{lr}
        \toprule
        Evaluation & Estimated cost (USD) \\
        \midrule
        Quality & 500.96 \\
        Local novelty & 139.17 \\
        Global novelty & 180.39 \\
        RINoBench evaluation & 37.41 \\
        \midrule
        \textbf{Total} & \textbf{857.93} \\
        \bottomrule
    \end{tabular}
    \caption{
        Estimated API costs for LLM-based evaluation,
        including development and validation runs.
        Global novelty includes web-search costs.
    }
    \label{tab:judge_api_cost}
\end{table}
\section{Additional Results Across Model Updates}
\label{app:temporal}

\begin{figure*}[t]
    \centering
    \includegraphics[width=0.95\linewidth]{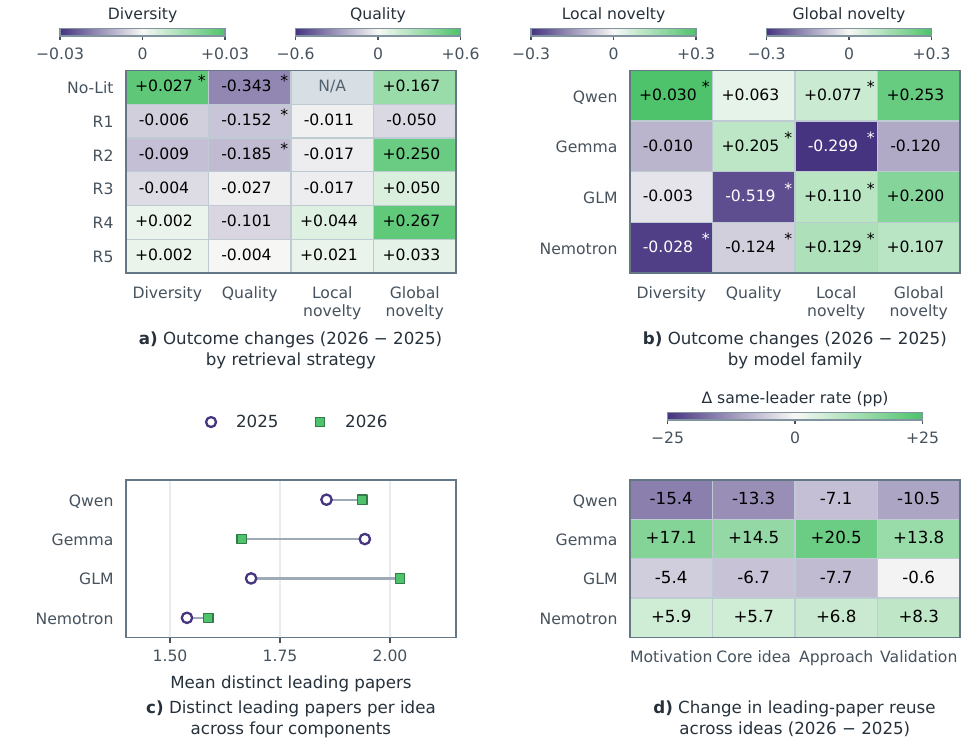}
    \caption{
        Comparison of matched 2025 and 2026 models.
        (a,b) Changes in diversity, quality, and novelty
        by retrieval strategy and model family.
        (c) Mean number of distinct leading papers across
        an idea's four components.
        (d) Change in leading-paper reuse for each component,
        measured among idea pairs generated from the same
        literature set, in percentage points.
        Heatmaps show 2026 minus 2025; positive values in
        (d) indicate greater reuse.
        Asterisks mark statistically significant changes.
    }
    \label{fig:model-updates}
\end{figure*}

Table~\ref{tab:temporal-retrieval-effects} extends the analysis in Section~\ref{sec:temporal-results} across successive model releases.
Figure~\ref{fig:model-updates} also reports changes in leading-paper counts and reuse.

\paragraph{Retrieval's diversity advantage narrows after 2025.}
For Gemma and Qwen, the mean diversity difference between retrieval and No Literature rises from $+0.016$ in 2024 to $+0.029$ in 2025, then falls to $-0.005$ in 2026 (Table~\ref{tab:temporal-retrieval-effects}).
The 2025--2026 decline also appears across all four families, whose mean difference changes from $+0.015$ to $-0.015$.
Over this update, diversity increases under No Literature while changing little under retrieval, reducing the additional variation supplied by retrieved papers.

\paragraph{The relative quality effect increases from 2025 to 2026.}
For Gemma and Qwen, the mean quality difference between retrieval and No Literature changes from $-0.129$ in 2024 to $-0.200$ in 2025 and approximately zero in 2026.
Across all four families, this difference increases from $-0.169$ to $+0.081$ between 2025 and 2026.
Mean quality falls by 0.094 under retrieval and by 0.343 under No Literature, yielding a 0.249 increase in retrieval's relative quality effect.
Retrieved literature therefore preserves more of the earlier models' quality as its diversity advantage declines.

\paragraph{The largest shifts occur in Qwen and GLM.}
From 2025 to 2026, Qwen and GLM have the largest estimated decreases in retrieval's diversity effect ($-0.049$ and $-0.060$) and the largest increases in its relative quality effect ($+0.396$ and $+0.415$).
Gemma's corresponding changes are $-0.017$ and $+0.005$, while Nemotron's are $+0.007$ and $+0.183$.

\begin{table*}[t]
    \centering
    \small
    \setlength{\tabcolsep}{5pt}
    \renewcommand{\arraystretch}{1.12}
    \begin{tabular*}{\textwidth}{@{\extracolsep{\fill}}lrrrrr@{}}
        \toprule
        & \multicolumn{3}{c}{Retrieval effect}
        & \multicolumn{2}{c}{Change} \\
        \cmidrule(lr){2-4}\cmidrule(l){5-6}
        Metric & 2024 & 2025 & 2026 & 2024--2025 & 2025--2026 \\
        \midrule
        \multicolumn{6}{@{}l}{\textit{Gemma and Qwen}} \\
        Diversity & $+0.016$ & $+0.029$ & $-0.005$ & $+0.012$ & $-0.033^{*}$ \\
        Quality & $-0.129$ & $-0.200$ & $0.000$ & $-0.071$ & $+0.200$ \\
        \midrule
        \multicolumn{6}{@{}l}{\textit{Gemma, Qwen, GLM, and Nemotron}} \\
        Diversity & --- & $+0.015$ & $-0.015$ & --- & $-0.030^{*}$ \\
        Quality & --- & $-0.169$ & $+0.081$ & --- & $+0.249^{*}$ \\
        \bottomrule
    \end{tabular*}
    \caption{Retrieval effects across model releases, averaged equally over R1--R5, the listed families, and 15 topics.
    Effects are retrieval minus No Literature; changes are later minus earlier.
    Asterisks mark changes with Benjamini--Hochberg adjusted $q<0.05$.}
    \label{tab:temporal-retrieval-effects}
\end{table*}

\paragraph{Changes in leading-paper reuse differ across families.}
Qwen's gains in diversity and local novelty accompany a 13.3-percentage-point decrease in core-idea leading-paper reuse (Figure~\ref{fig:model-updates}c,d).
Gemma's quality gain accompanies an increase in core-idea reuse of 14.5 percentage points, while its mean number of distinct leading papers per idea falls from 1.94 to 1.66.
GLM shows less reuse and more leading papers per idea, alongside higher local novelty.
Qwen and Nemotron show little change in their leading-paper counts, despite opposite changes in both reuse and diversity: reuse decreases and diversity increases for Qwen, whereas reuse increases and diversity decreases for Nemotron.

\section{Detailed results on diversity, quality, and semantic distance}
\label{app:rq1-dimension-results}

We examine diversity across idea components, quality scores by
dimension, and semantic distances to the input literature across
the four 2025 models and 15 research topics.
\cnolit denotes generation without retrieved literature.
R1--R5 denote Semantic Top-$K$, Most-Cited Top-$K$,
Semantic + Entity + Co-citation, Core-Paper Reference Expansion,
and Chain of Ideas, respectively.

\subsection{Diversity across idea components}
\label{app:rq1-component-diversity}

\begin{figure*}[t]
    \centering
    \includegraphics[width=\linewidth]{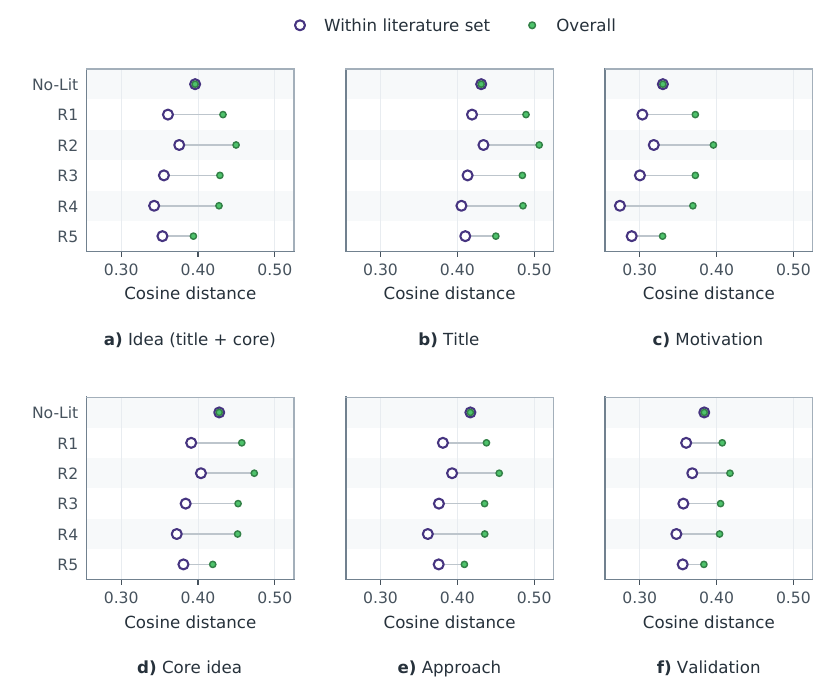}
    \caption{Semantic diversity across idea components, measured
    by mean pairwise cosine distance. Open circles show within-set
    diversity among ideas generated from the same papers.
    Filled circles show overall diversity among ideas from all
    literature sets for the same strategy, model, and topic.
    Panel (a) uses the title and core idea together;
    panels (b)--(f) use individual components.
    All panels share the same distance scale.}
    \label{fig:app-rq1-diversity}
\end{figure*}

\paragraph{Overall diversity exceeds within-set diversity in every component.}
This pattern holds for all five retrieval strategies across titles,
motivations, core ideas, approaches, and validation plans
(Figure~\ref{fig:app-rq1-diversity}).
R4 has the largest gap between the two measures: 0.084 when using
the title and core idea together, and 0.056--0.095 across individual
components. For motivations, its within-set and overall diversity
are 0.274 and 0.369, respectively.

\paragraph{R2 produces the highest overall diversity across components.}
R2 ranks first for titles (0.507), motivations (0.396),
core ideas (0.473), approaches (0.455), and validation plans (0.417).
R5 exceeds \cnolit in overall title diversity
(0.450 vs.\ 0.431), while its other four components remain
at or below their \cnolit means.

\subsection{Quality across evaluation dimensions}
\label{app:rq1-quality-dimensions}

\begin{figure*}[t]
    \centering
    \includegraphics[width=\linewidth]{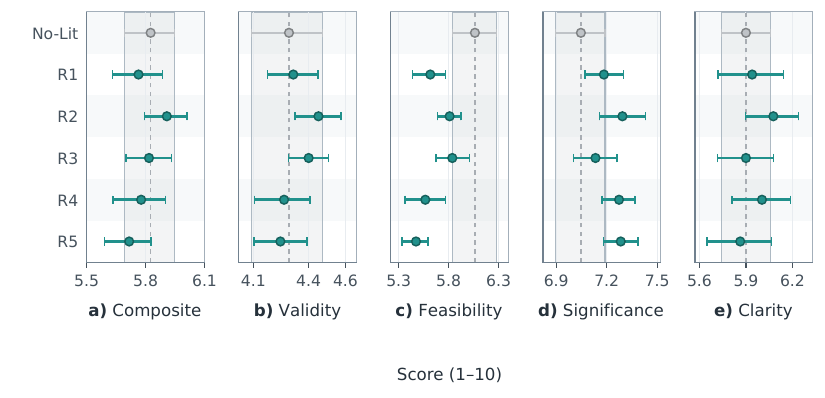}
    \caption{Quality scores by dimension. The composite is
    the mean of validity, feasibility, significance, and clarity.
    Points show mean scores, with 95\% confidence intervals
    obtained by bootstrapping topics. Shaded bands and dashed
    lines indicate the \cnolit intervals and means, respectively.
    Horizontal axis ranges differ across panels.}
    \label{fig:app-rq1-quality}
\end{figure*}

\paragraph{Significance scores are higher and feasibility scores are lower with retrieval.}
Every retrieval strategy has a higher mean significance score
than \cnolit, increasing from 7.05 to 7.13--7.29,
while mean feasibility falls from 6.06 to 5.47--5.84
(Figure~\ref{fig:app-rq1-quality}).
R5 shows this pattern most clearly: its mean significance
is 7.28, while its feasibility is 5.47, the lowest
among the six conditions.

\paragraph{R2 has the highest mean composite quality.}
R2 achieves a composite score of 5.91, compared with
5.83 for \cnolit. It also has the highest mean validity
(4.45), significance (7.29), and clarity (6.08).
Mean composite scores range from 5.72 to 5.91
across retrieval strategies.

\subsection{Semantic distance to the input literature}
\label{app:rq1-literature-distance}

\begin{figure*}[t]
    \centering
    \includegraphics[width=\linewidth]{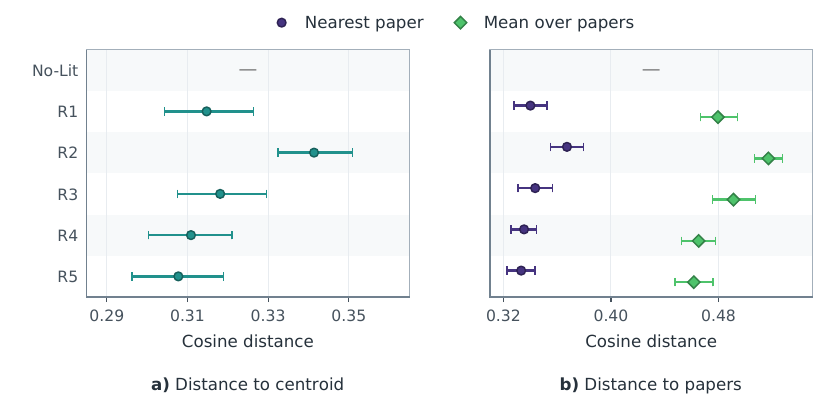}
    \caption{Semantic distance from ideas to their input literature.
    Panel (a) shows cosine distance to the centroid of the eight
    input-paper embeddings. Panel (b) shows distance to the nearest
    paper (circles) and mean distance across the input papers
    (diamonds). Points show means, with 95\% confidence intervals
    obtained by bootstrapping topics. Dashes denote undefined
    distances for \cnolit.}
    \label{fig:app-rq1-novelty}
\end{figure*}

\paragraph{R2 yields the largest distances across all three measures.}
The three measures give the same strategy ranking:
R2, R3, R1, R4, and R5
(Figure~\ref{fig:app-rq1-novelty}).
Under R2, mean distances are 0.341 to the literature-set
centroid, 0.367 to the nearest paper, and 0.517 when averaged
over the input papers. R5 has the smallest corresponding
distances: 0.308, 0.333, and 0.462.

\section{Results of closed-source models}
\label{app:closed-models-retrieval}

Table~\ref{tab:detailed-results-2026} reports quality, diversity, and novelty for the three closed-source models in the 2026 evaluation under no-lit and R1--R5. GPT-5.6 Luna achieves the highest aggregate quality under every strategy (6.94--7.17). GPT-5.6 Luna and Gemini~3.5 Flash-Lite attain their highest aggregate quality under no-lit (7.17 and 6.33), while Claude Haiku~4.5 performs best under R4 (6.33).

R3 yields the highest local novelty for all three models, tying with R4 for Claude Haiku~4.5. The highest overall diversity occurs under R4 for GPT-5.6 Luna (0.342), R1 for Gemini~3.5 Flash-Lite (0.400), and R2 for Claude Haiku~4.5 (0.378). Thus, the strategies associated with the highest quality, diversity, and novelty differ across models and metrics.

\begin{table*}[!htbp]
\centering
\small
\setlength{\tabcolsep}{3pt}
\renewcommand{\arraystretch}{1.0}
\begin{tabular*}{\linewidth}{@{\extracolsep{\fill}}lrrrrrrrrr@{}}
\toprule
& \multicolumn{5}{c}{Quality} & \multicolumn{2}{c}{Diversity} & \multicolumn{2}{c}{Novelty} \\
\cmidrule(lr){2-6}\cmidrule(lr){7-8}\cmidrule(l){9-10}
Strategy & V & F & S & C & Avg. & W & O & L & G \\
\midrule
\multicolumn{10}{@{}l}{\textit{GPT-5.6 Luna}} \\
no-lit & 6.43 & \textbf{6.32} & 7.97 & 7.97 & \textbf{7.17} & \textbf{0.273} & 0.273 & --- & 2.67 \\
R1 & 6.41 & 5.80 & 8.04 & 7.87 & 7.03 & 0.230 & 0.328 & 3.65 & 2.53 \\
R2 & 6.40 & 5.97 & 8.01 & 7.97 & 7.09 & 0.229 & 0.339 & 3.79 & 2.67 \\
R3 & 6.33 & 5.84 & 8.05 & 7.97 & 7.05 & 0.223 & 0.224 & \textbf{3.84} & \textbf{2.87} \\
R4 & \textbf{6.45} & 6.00 & 8.01 & \textbf{8.05} & 7.13 & 0.215 & \textbf{0.342} & 3.69 & \textbf{2.87} \\
R5 & 6.16 & 5.69 & \textbf{8.07} & 7.83 & 6.94 & 0.226 & 0.308 & 3.68 & 2.73 \\
\midrule
\multicolumn{10}{@{}l}{\textit{Gemini 3.5 Flash-Lite}} \\
no-lit & 4.80 & \textbf{6.29} & 7.40 & \textbf{6.84} & \textbf{6.33} & \textbf{0.314} & 0.314 & --- & 2.33 \\
R1 & 4.76 & 5.99 & 7.55 & 6.48 & 6.19 & 0.261 & \textbf{0.400} & 3.37 & 2.47 \\
R2 & 4.84 & 5.72 & 7.56 & 6.61 & 6.18 & 0.261 & 0.398 & 3.37 & \textbf{2.57}$^{\dagger}$ \\
R3 & \textbf{4.85} & 5.71 & \textbf{7.65} & 6.60 & 6.20 & 0.235 & 0.237 & \textbf{3.43} & 2.27 \\
R4 & 4.75 & 5.79 & 7.45 & 6.37 & 6.09 & 0.239 & 0.391 & 3.09 & 2.27 \\
R5 & 4.61 & 5.52 & 7.51 & 6.31 & 5.99 & 0.226 & 0.359 & 3.09 & 2.40 \\
\midrule
\multicolumn{10}{@{}l}{\textit{Claude Haiku 4.5}} \\
no-lit & 4.88 & 5.53 & 7.68 & 6.92 & 6.25 & \textbf{0.347} & 0.347 & --- & 2.20 \\
R1 & 4.81 & 5.27 & 7.69 & 6.93 & 6.18 & 0.278 & 0.364 & 3.43 & 2.33 \\
R2 & 4.96 & 5.15 & \textbf{7.87} & 7.01 & 6.25 & 0.279 & \textbf{0.378} & 3.39 & 2.53 \\
R3 & 4.69 & 5.37 & 7.81 & 6.87 & 6.19 & 0.280 & 0.281 & \textbf{3.49} & \textbf{2.60} \\
R4 & \textbf{4.97} & \textbf{5.59} & 7.68 & \textbf{7.07} & \textbf{6.33} & 0.266 & 0.371 & \textbf{3.49} & 2.33 \\
R5 & 4.89 & 5.29 & 7.85 & 7.05 & 6.27 & 0.276 & 0.332 & 3.37 & 2.27 \\
\bottomrule
\end{tabular*}
\caption{Detailed results for the three closed-source models in the 2026 evaluation. V: Validity; F: Feasibility; S: Significance; C: Clarity; Avg.: aggregate quality. W: within-set diversity; O: overall diversity. L: local novelty; G: global novelty. Bold marks the highest displayed value within each model and column. Local novelty summarizes 75 ideas per model and retrieval strategy; it is not reported for no-lit. $^{\dagger}$Global novelty averages 14 topics for this entry (15 otherwise).}
\label{tab:detailed-results-2026}
\end{table*}

\section{Reasoning-Enabled Ideation}
\label{app:reasoning_ideation}

\subsection{Setup}
\label{app:reasoning_setup}

We compare the four 2026 models with and without reasoning, using
matched topics, literature sets, and prompts. Both modes use $T=1$;
reasoning uses top-$p=0.95$ and an 8,192-token cap, and the baseline
uses default top-$p$ and a 3,600-token cap. Evaluation samples are
paired by model, topic, condition, and sample index
(Table~\ref{tab:reasoning_coverage}). Local novelty uses a balanced subset comprising the first two
literature sets and five ideas per set. Global novelty uses the first four jointly evaluated sample
indices per cell for the first two topics in Table~2. Metrics and
judges follow Section~\ref{sec:evaluation}, with title and core-idea embeddings as in
Appendix~E.1. Means weight models and topics equally; confidence
intervals use 10,000 paired topic-bootstrap resamples.

\begin{table}[!ht]
\centering
\small
\setlength{\tabcolsep}{3.5pt}
\begin{tabular}{lrrr}
\toprule
Analysis & Topics & Samples/cell & Total/mode \\
\midrule
Diversity & 15 & 50 & 18,000 \\
Quality & 15 & 5 & 1,800 \\
Local novelty & 15 & 10 & 3,000 \\
Global novelty & 2 & 4 & 192 \\
\bottomrule
\end{tabular}
\caption{Evaluation samples per mode. A cell is a
model--topic--condition combination. Local novelty covers R1--R5;
the other analyses also include No Literature.}
\label{tab:reasoning_coverage}
\end{table}

\begin{figure*}[t]
\centering
\includegraphics[width=\textwidth]{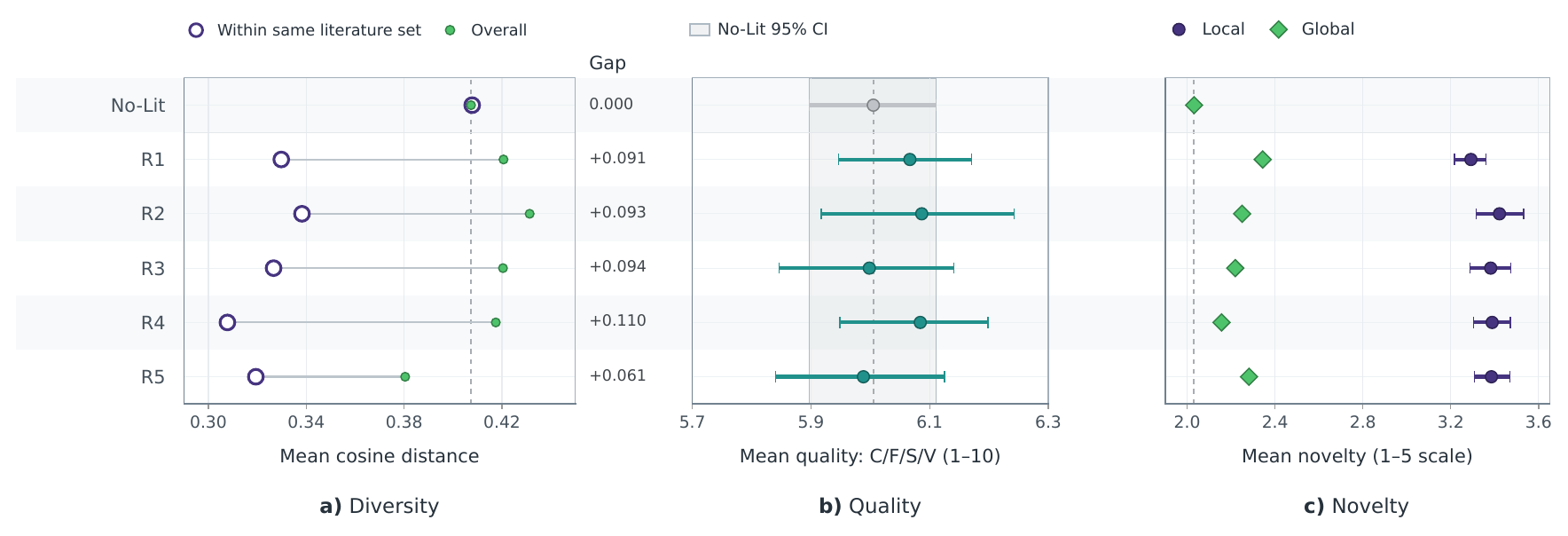}
\caption{Reasoning-enabled ideation across four models.
(a) Diversity within literature sets and overall, using title and
core-idea embeddings. Gap denotes overall minus within-set diversity.
(b) Quality; the shaded band shows the No-Lit 95\% confidence interval.
(c) Local novelty across 15 topics and global novelty across two topics.
Quality and local-novelty error bars are 95\% topic-bootstrap confidence
intervals.}
\label{fig:reasoning_diversity}
\end{figure*}

\begin{table*}[t]
\centering
\small
\setlength{\tabcolsep}{4.5pt}
\begin{tabular}{lrrrrrrrrr}
\toprule
& \multicolumn{3}{c}{Quality (1--10)}
& \multicolumn{3}{c}{Local novelty (1--5)}
& \multicolumn{3}{c}{Global novelty (1--5)} \\
\cmidrule(lr){2-4}\cmidrule(lr){5-7}\cmidrule(lr){8-10}
Model & Off & On & $\Delta$ & Off & On & $\Delta$ & Off & On & $\Delta$ \\
\midrule
Qwen3.6 & 5.792 & 6.225 & $+0.433$ & 3.264 & 3.541 & $+0.277$ & 2.208 & 2.458 & $+0.250$ \\
Gemma 4 & 6.280 & 6.448 & $+0.168$ & 3.235 & 3.513 & $+0.279$ & 2.250 & 1.979 & $-0.271$ \\
Nemotron Cascade 2 & 5.498 & 5.690 & $+0.192$ & 2.929 & 3.043 & $+0.113$ & 2.292 & 2.188 & $-0.104$ \\
GLM-4.7-Flash & 5.272 & 5.790 & $+0.518$ & 3.229 & 3.397 & $+0.168$ & 2.396 & 2.229 & $-0.167$ \\
Macro-average & 5.711 & 6.038 & $+0.328$ & 3.164 & 3.374 & $+0.209$ & 2.286 & 2.214 & $-0.073$ \\
\bottomrule
\end{tabular}
\caption{Quality and novelty with reasoning disabled (Off) and enabled
(On); $\Delta=\mathrm{On}-\mathrm{Off}$. Each model contributes
450 quality, 750 local-novelty, and 48 global-novelty paired samples.
Quality and global novelty average over six conditions; local novelty
over R1--R5. Global novelty covers two topics; the other scores cover 15.}
\label{tab:reasoning_scores}
\end{table*}

\subsection{Results and Implications}
\label{app:reasoning_results}

\paragraph{Reasoning does not remove the need to vary the literature.}
With reasoning enabled, all five retrieval strategies reduce within-set
diversity relative to No Literature by 17.1\%--24.5\%
(Figure~\ref{fig:reasoning_diversity}a). Pooling ideas from different
literature sets gives R1--R4 higher mean diversity than No Literature
by 2.5\%--5.9\%; R5 remains 6.6\% below it.
The gap between overall and within-set diversity increases under
\emph{every} retrieval strategy when reasoning is enabled, from
0.077 to 0.090 on average across R1--R5. R4 retains the largest gap
(0.101 without reasoning; 0.110 with reasoning).
This separation also holds across all five individual idea fields,
with gaps of 0.039--0.119 under reasoning.
Thus, the concentration of ideas around a fixed literature context
persists even when the model reasons before answering. For an ideation
pipeline seeking broader exploration, enabling reasoning leaves
literature selection as an important design choice: different paper
sets provide variation that repeated generation from the same papers
does not recover.

\paragraph{Better individual proposals need not form a more diverse collection.}
Quality increases for all four models
(Table~\ref{tab:reasoning_scores}), with the mean rising from 5.711
to 6.038 ($\Delta=+0.328$; 95\% CI $[0.291,0.365]$).
Clarity contributes the largest improvement ($+0.633$), followed by
validity ($+0.314$), significance ($+0.216$), and feasibility
($+0.148$). Yet mean overall diversity falls from 0.424 to 0.413
across the six conditions. This contrast appears in three of the four
models: Qwen, GLM, and Nemotron gain quality while losing diversity.
Gemma gains both, so the trade-off is not universal.
The observation separates proposal development from exploration:
higher-rated individual ideas can still cover a narrower range of
directions. Reasoning should therefore be evaluated using both
individual-idea scores and collection-level diversity; an increase in
quality alone does not establish a broader scientific search.

\paragraph{Local novelty gains do not establish novelty against the broader literature.}
Local novelty increases for all four models, from 3.164 to 3.374 on
average ($\Delta=+0.209$; 95\% CI $[0.170,0.246]$).
The two-topic search-assisted global-novelty evaluation shows a
different pattern: the mean changes from 2.286 to 2.214, with an
increase for Qwen ($+0.250$) and decreases for Gemma ($-0.271$),
Nemotron ($-0.104$), and GLM ($-0.167$).
The local result shows greater novelty relative to the supplied
papers. 

\section{Retrieval Effects Across Topics}
\label{app:topic-retrieval-preferences}

\begin{figure*}[p]
    \centering
    \includegraphics[
        width=\textwidth,
        height=0.80\textheight,
        keepaspectratio
    ]{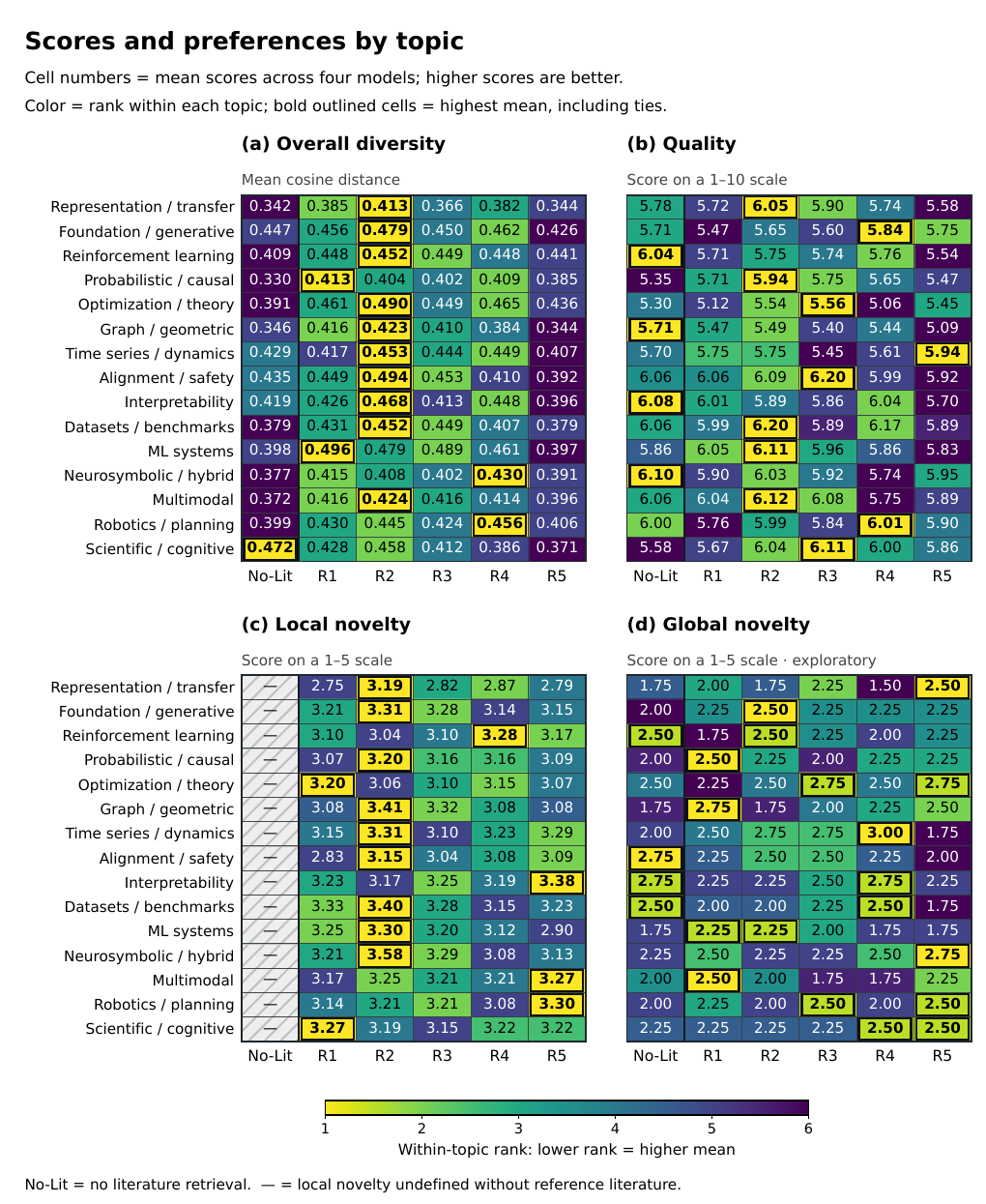}
    \caption{
        Topic-level scores for \cnolit{} and the five retrieval
        strategies, averaged across the four 2025 models.
        Cell numbers show mean scores; colors indicate ranks
        within each topic, with lower ranks indicating higher
        scores.
        Bold outlines mark the highest means, including ties.
        Rankings use unrounded scores and average ranks for ties.
        Gray \cnolit{} cells in panel (c) indicate that local
        novelty is undefined without reference literature.
    }
    \label{fig:topic-strategy-preferences}
\end{figure*}

\begin{figure*}[p]
    \centering
    \includegraphics[
        width=\textwidth,
        height=0.80\textheight,
        keepaspectratio
    ]{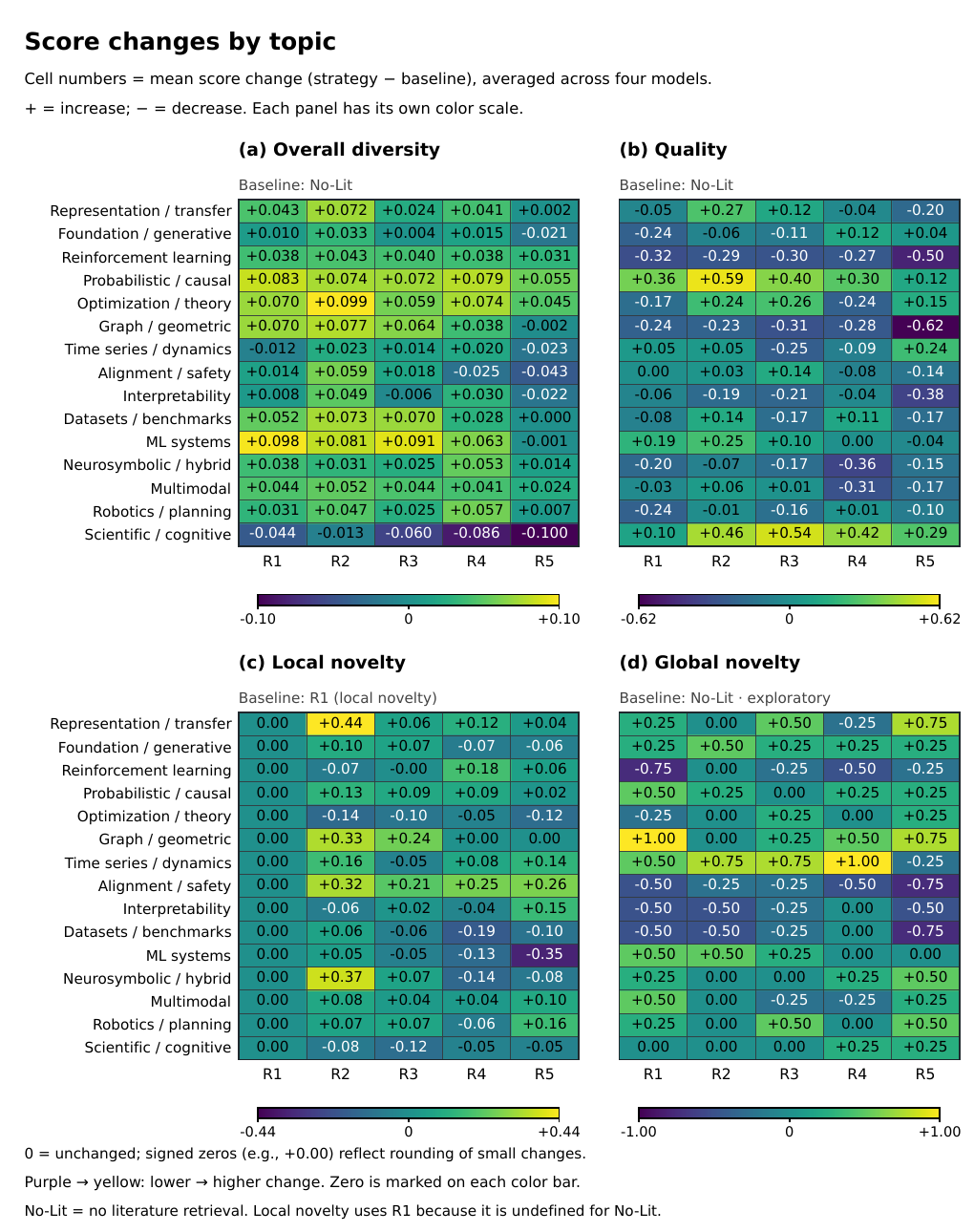}
    \caption{
        Topic-level changes in diversity, quality, and novelty,
        averaged across the four 2025 models.
        Values show differences from \cnolit{} in panels
        (a), (b), and (d), and from R1 in panel (c).
        Positive values indicate increases.
        Each panel uses a separate color scale symmetric
        around zero.
    }
    \label{fig:topic-strategy-effects}
\end{figure*}

R2 achieves the highest mean diversity in 10 of the 15 topics
and the highest mean quality in five, with \cnolit{} included
in both comparisons.
Figures~\ref{fig:topic-strategy-preferences}
and~\ref{fig:topic-strategy-effects} report topic-level scores
and changes relative to the corresponding baselines,
averaged across the four 2025 models.

\paragraph{Diversity--quality trade-offs differ across topics.}
In scientific and cognitive applications, all five retrieval
strategies reduce diversity and increase quality relative
to \cnolit{}.
R3 achieves the highest quality, with a gain of $0.538$
and an increase in feasibility of $1.00$ point.
In reinforcement learning, all five strategies increase
diversity and reduce quality, with feasibility decreasing
by $0.80$--$1.25$ points.
R4 achieves the highest local novelty and quality among
the retrieval strategies in this topic, but its quality
score remains $0.275$ below \cnolit{}.

\paragraph{Local and global novelty yield different strategy rankings.}
In representation, transfer, meta, and continual learning,
R2 scores higher than R5 on local novelty
($3.190$ versus $2.795$).
The ordering reverses in the exploratory global novelty
evaluation, where R5 scores $2.50$ and R2 scores $1.75$.
The two metrics use different references: local novelty
compares each idea with its supplied literature set,
while global novelty uses literature found through
web search.
R2 also has higher quality than R5 in this topic
($6.05$ versus $5.58$).

\section{Checking Paper Attribution}
\label{app:attribution_robustness}

We examine whether these attribution scores provide meaningful information about the relationship between retrieved papers and generated ideas, using human judgments and model-reported sources as complementary evidence. We also examine disagreements and discuss why evaluating attribution requires distinguishing different kinds of paper contributions.

\paragraph{How we compute attribution.}
For each generated idea, we remove one input paper at a time and measure how the model's predictions change when scoring the same idea text. Larger changes give higher attribution scores. We compute scores from content tokens and express them as percentages that sum to 100\% across the eight input papers within each of the idea's four parts.

\subsection{Comparison with Human Judgments}
\label{app:attr_human}

The Spearman correlation between attribution scores and human ratings is 0.454. For comparison, the correlation between human raters is 0.360, computed on the same samples using the same procedure. Agreement is therefore limited, including among human raters.

These results motivate a finer-grained classification of what each paper contributes. Our scores distinguish papers and idea parts, but do not explicitly distinguish contribution types: a paper may motivate a problem, supply a method, or inform evaluation. Human raters may also weigh these roles differently. Evaluating these contributions separately could help explain the disagreement.

\subsection{Reported Sources Usually Include the Highest-scoring Paper}
\label{app:attr_selfreport}

We also compare attribution with sources reported by the model, either after it has generated an idea or while it is generating one. For each idea part, we check whether the reported sources include the paper with the highest attribution score.

When sources are requested after generation, this paper is reported in 77.58\% of idea parts. Using another idea's source report instead lowers this proportion to 52.97\% (Table~\ref{tab:attr_validation}). The comparison uses ideas generated from the same eight input papers and matches corresponding parts.

The 24.61-percentage-point difference shows that the correspondence depends on the specific idea, beyond a tendency to report the same papers across ideas. The difference is positive for all four idea parts. Requesting sources during generation produces a similar pattern: 71.89\% with the idea's own report, compared with 48.36\% with another idea's report.

\begin{table}[!t]
\centering
\small
\setlength{\tabcolsep}{5pt}
\begin{tabular}{@{}lrr@{}}
\toprule
\shortstack[l]{Sources\\requested}
& \shortstack{This idea's\\report}
& \shortstack{Another idea's\\report} \\
\midrule
After generation  & 77.58\% & 52.97\% \\
During generation & 71.89\% & 48.36\% \\
\bottomrule
\end{tabular}
\caption{
Percentage of idea parts whose highest-scoring paper appears in the source report. Each row covers 11,200 idea parts.
The comparison with another idea averages over the corresponding part of the other 19 ideas generated from the same inputs. Empty reports count as no match.
In 20,000 shuffles of whole source reports among ideas within each setting, none reached the observed correspondence in either condition (Holm-adjusted $p\approx0.00010$ across the two conditions). }
\label{tab:attr_validation}
\end{table}

\subsection{Why Attribution and Reported Sources Can Disagree}
\label{app:attr_cases}

\begin{figure*}[!t]
\centering
\includegraphics[width=0.94\textwidth]{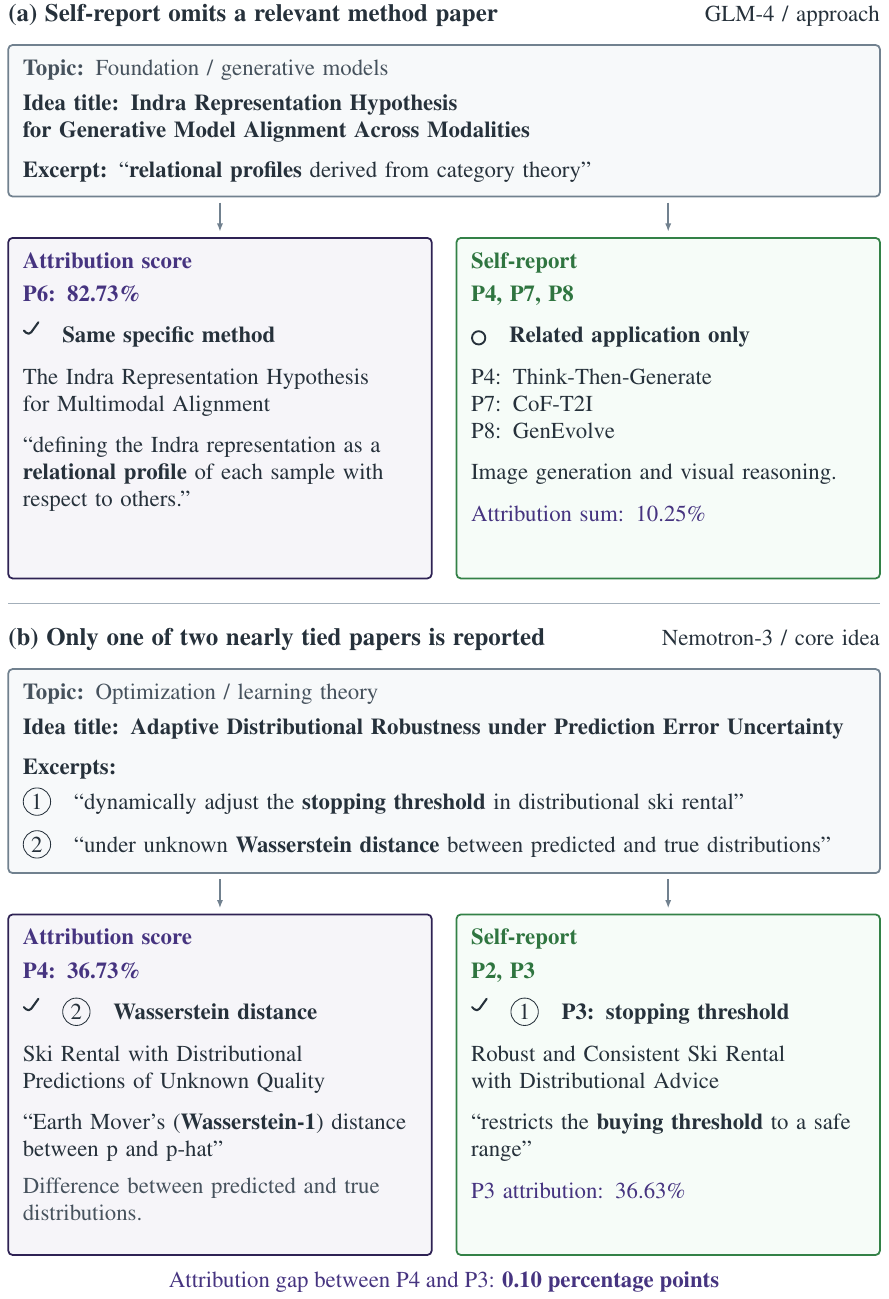}
\caption{
Two examples of disagreement between attribution and reported
sources: (a) a relevant method paper is omitted;
(b) one of two almost equally scored papers is reported.
Percentages show attribution scores normalized across eight papers.
Quotes reproduce saved text; paper numbers are local to each case.
}
\label{fig:attr_cases}
\end{figure*}

\paragraph{A relevant method paper can be omitted.}
In Figure~\ref{fig:attr_cases}a, GLM-4 proposes using Indra representations and relationships between samples to align outputs across modalities. P6, \emph{The Indra Representation Hypothesis for Multimodal Alignment}, describes this method and receives 82.73\% of the attribution score.
Yet the model reports P4, P7, and P8, which concern image generation and visual reasoning and together receive only 10.25\%. 
This case suggests that attribution can identify a method-level connection missed by the model's source report.

\paragraph{Almost equal scores can count as a disagreement.}
In Figure~\ref{fig:attr_cases}b, Nemotron-3 proposes adjusting a decision threshold as prediction error is estimated. The idea combines elements of P3, which studies robust threshold selection, and P4, which studies errors between predicted and actual distributions. P4 receives 36.73\% of the score and P3 receives 36.63\%.
The model reports P2 and P3, but not P4. Our comparison therefore counts this as a disagreement, even though the reported P3 scores only \textbf{0.10 percentage points} below P4.

These examples show why checking only the highest-scoring paper gives an incomplete picture. Understanding a disagreement also requires examining what the papers contribute and how far apart their scores are.

\subsection{Generate Ideas with Citations}
\label{app:attr_protocol}

We also examine whether requesting sources during generation
changes the ideas themselves (Table~\ref{tab:attr_generation}).
The average number of distinct highest-scoring papers across an
idea's four parts increases from 1.750 to 2.014.
Ideas generated from the same inputs also become more similar:
their average distance falls by 10.02\%.
However, neither difference is statistically significant across
the five topics (both Holm-adjusted $p=0.375$).

\begin{table}[!t]
\centering
\small
\setlength{\tabcolsep}{4pt}
\begin{tabular}{@{}lrr@{}}
\toprule
& \multicolumn{2}{c}{Sources requested} \\
\cmidrule(l){2-3}
Measure
& \shortstack{After\\generation}
& \shortstack{During\\generation} \\
\midrule
\shortstack[l]{Distinct highest-scoring\\papers per idea}
& 1.750 & 2.014 \\
Average distance between ideas
& 0.3505 & 0.3154 \\
\bottomrule
\end{tabular}
\caption{
Comparison of 2,800 ideas per condition.
The first row counts distinct papers ranked first across an
idea's four parts. The second reports the average cosine distance
between embeddings of titles and core ideas generated from
the same inputs; lower values indicate more similar ideas.
}
\label{tab:attr_generation}
\end{table}
\section{Licenses and Artifact Release}
\label{app:licenses}

The Qwen checkpoints and Gemma 4 used in this study are distributed
under the Apache License 2.0, and the GLM checkpoints under the MIT
License. Gemma 2 and Gemma 3 are distributed under the Gemma Terms
of Use. Nemotron 3 Nano and Nemotron Cascade 2 are distributed under
the NVIDIA Nemotron Open Model License and the NVIDIA Open Model
License, respectively.

Paper titles and abstracts were obtained from arXiv, whose metadata
are available under the CC0 1.0 Universal Public Domain
Dedication.\footnote{
\url{https://info.arxiv.org/help/license/index.html}}
This metadata license does not extend to article full texts, which
remain subject to their individual licenses.
Citation metadata were obtained from Semantic
Scholar~\citep{kinney2023semantic}. Access to its API is governed by
the Semantic Scholar API License Agreement, while the supplied data
and any underlying third-party content remain subject to their
applicable licenses.\footnote{
\url{https://www.semanticscholar.org/product/api/license}}
\section{A Case Study of Literature Attribution}
\label{app:source_case_study}

\begin{figure*}[t]
    \centering
    \includegraphics[width=\linewidth]{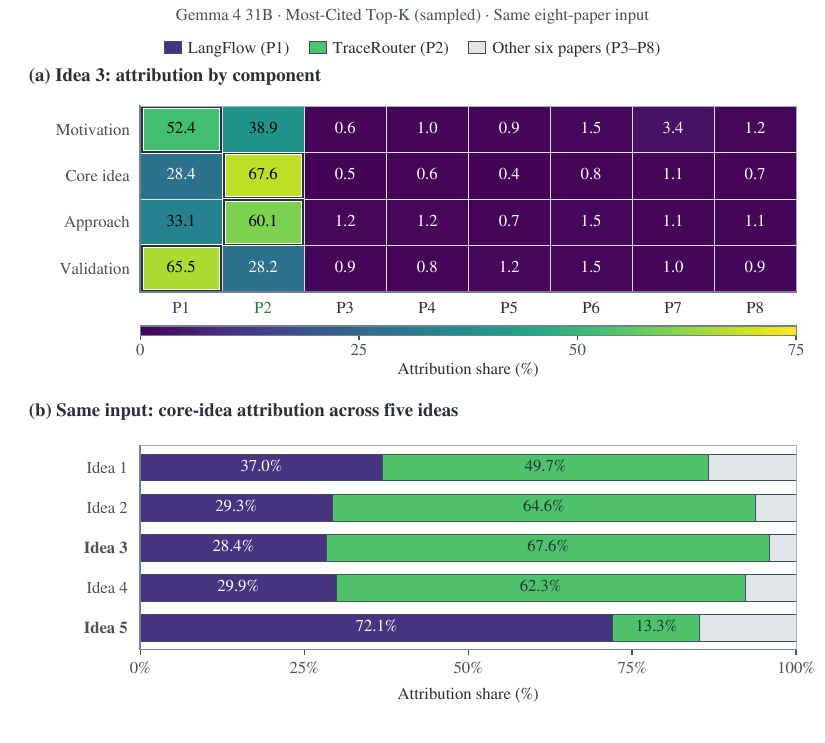}
\caption{
    Component-level attribution for ideas generated from
    the same literature set.
    (a) Attribution shares for Idea 3's four components
    across all eight papers. Color encodes attribution
    share; thicker outlines mark the largest share
    in each component.
    (b) Core-idea attribution shares across five ideas,
    with LangFlow in purple, TraceRouter in green,
    and the other six papers aggregated in gray.
    TraceRouter leads Ideas 1--4, while LangFlow leads
    Idea 5.
    P1 and P2 denote LangFlow and TraceRouter in the
    input order.
}
    \label{fig:rq2-nlp-case}
\end{figure*}

\begin{figure*}[p]
    \centering
    \includegraphics[
        width=\linewidth,
        height=0.82\textheight,
        keepaspectratio
    ]{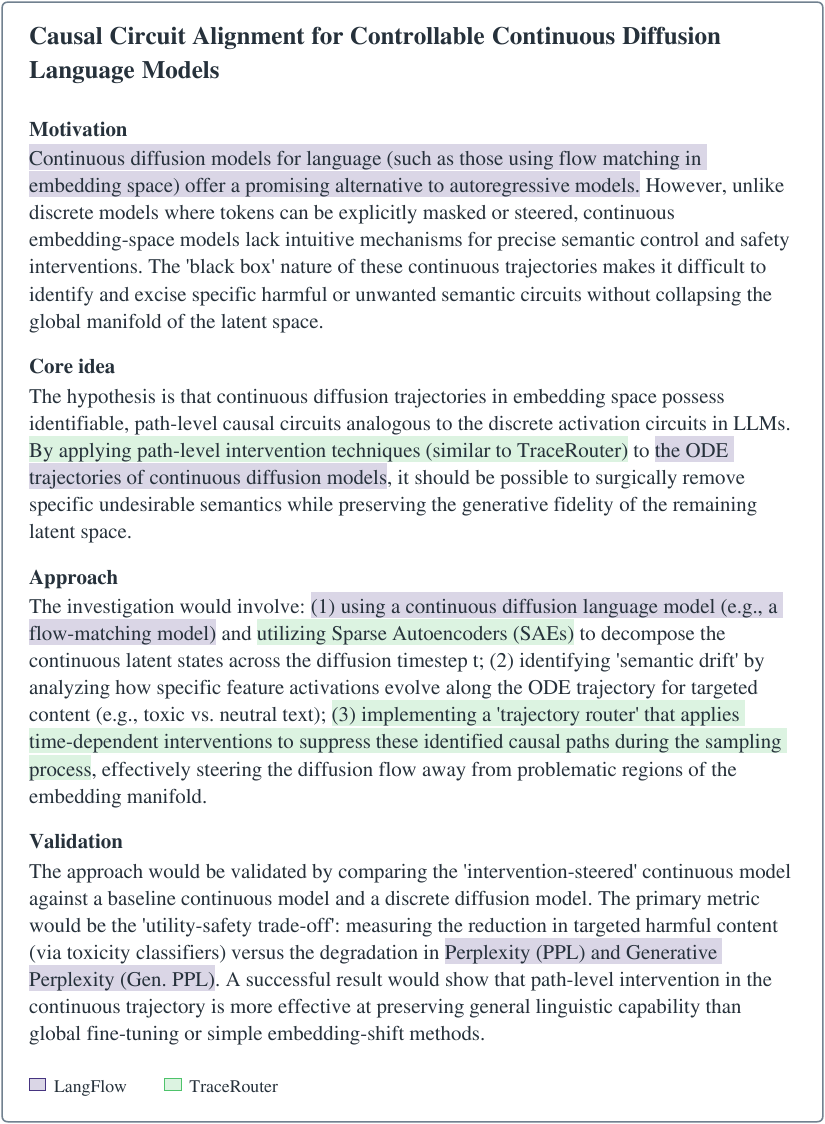}
\caption{
    Idea 3 with manual highlights.
    Purple passages mark correspondences with LangFlow's
    continuous-generation setting and evaluation metrics.
    Green passages mark correspondences with TraceRouter's
    path-level intervention and feature analysis using
    sparse autoencoders.
}
    \label{fig:rq2-idea3-card}
\end{figure*}

\begin{figure*}[p]
    \centering
    \includegraphics[
        width=\linewidth,
        height=0.82\textheight,
        keepaspectratio
    ]{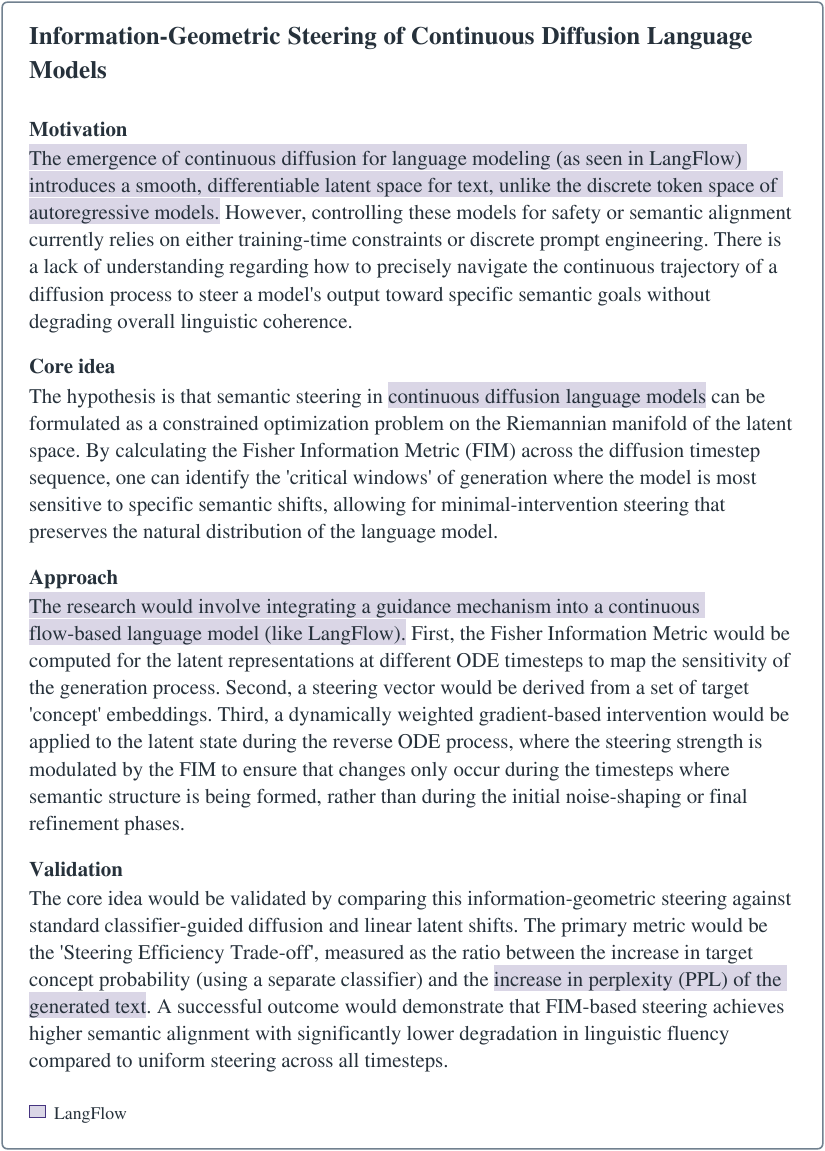}
\caption{
    Idea 5 with manual highlights.
    Purple passages mark correspondences with LangFlow's
    continuous-generation setting and perplexity-based
    evaluation.
}
    \label{fig:rq2-idea5-card}
\end{figure*}

\begin{figure*}[p]
    \centering
    \includegraphics[
        width=\linewidth,
        height=0.3\textheight,
        keepaspectratio
    ]{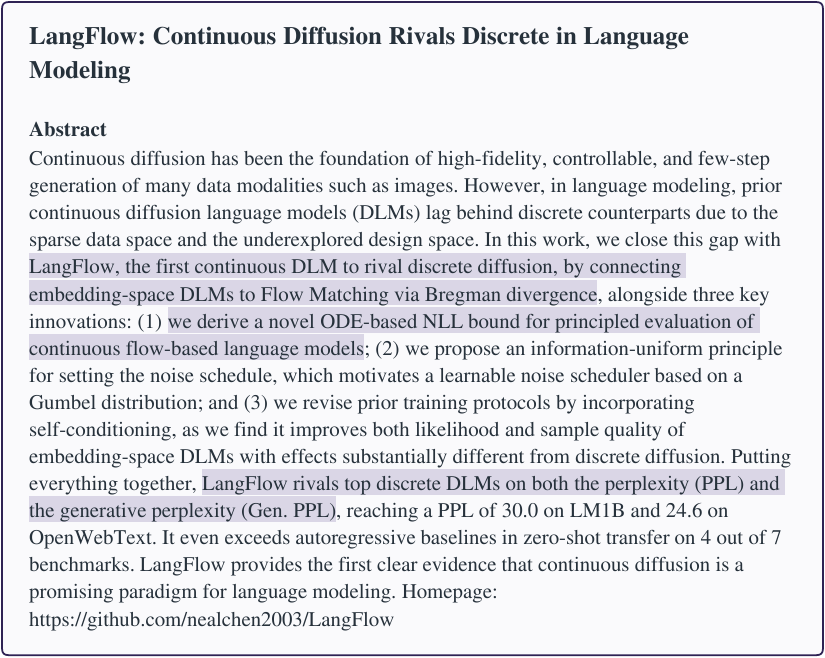}
\caption{
    LangFlow (P1): title and abstract supplied to the model.
    Purple highlights mark passages on continuous diffusion,
    flow matching, ODE-based evaluation, and perplexity
    metrics.
}

    \label{fig:rq2-langflow-card}
\end{figure*}

\begin{figure*}[p]
    \centering
    \includegraphics[
        width=\linewidth,
        height=0.3\textheight,
        keepaspectratio
    ]{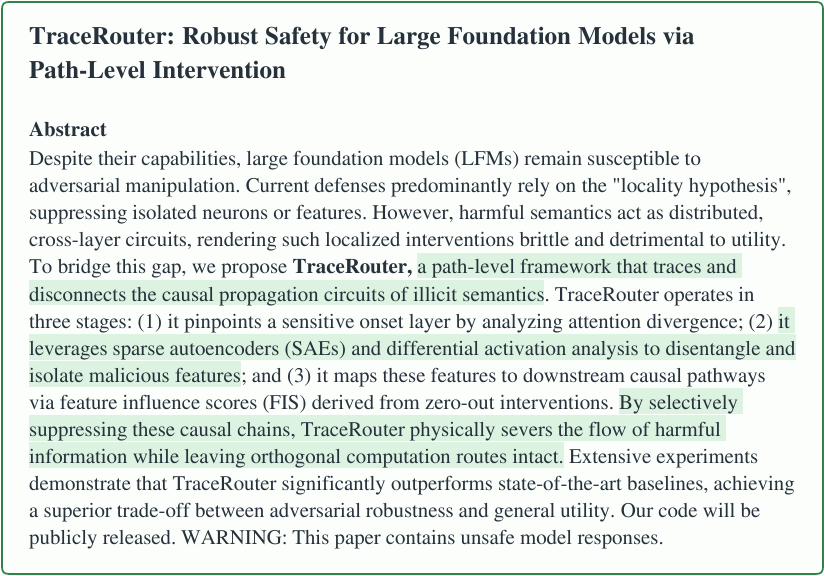}
\caption{
    TraceRouter (P2): title and abstract supplied to the model.
    Green highlights mark passages on path-level intervention,
    feature analysis using sparse autoencoders, and selective
    suppression of harmful causal chains.
}
    \label{fig:rq2-tracerouter-card}
\end{figure*}

We present five ideas generated by Gemma 4 (31B) from the
same eight-paper literature set using Most-Cited Top-K.
Figure~\ref{fig:rq2-nlp-case} shows their component-level
attribution.
We examine Ideas 3 and 5 in detail;
Figures~\ref{fig:rq2-idea3-card}
and~\ref{fig:rq2-idea5-card} show their complete text.
Figures~\ref{fig:rq2-langflow-card}
and~\ref{fig:rq2-tracerouter-card} reproduce the input
abstracts of LangFlow~\citep{chen2026langflow}
and TraceRouter~\citep{shi2026tracerouter}.

\paragraph{Manual annotation.}
Blue and gold highlights mark manually identified textual
correspondences with LangFlow and TraceRouter, respectively.

\paragraph{Different papers lead different components of Idea 3.}
Idea 3 proposes path-level control of continuous diffusion
language generation.
Its core idea explicitly connects
``path-level intervention techniques'' to TraceRouter
and applies them to ``ODE trajectories of continuous
diffusion models.''
The approach proposes combining a flow-matching model,
sparse autoencoders, and a trajectory router to suppress
selected paths during sampling.

TraceRouter leads the core idea ($67.6\%$) and approach
($60.1\%$), while LangFlow leads motivation ($52.4\%$)
and validation ($65.5\%$).
The highlighted passages connect the intervention method
to TraceRouter and the continuous-generation setting
to LangFlow.
The validation component also names perplexity and
generative perplexity, both reported in the LangFlow
abstract.

\paragraph{The leading paper differs across ideas.}
TraceRouter leads the core idea in Ideas 1--4.
In Idea 5, LangFlow accounts for $72.1\%$ of the core-idea
attribution, and TraceRouter accounts for $13.3\%$.

Idea 5 proposes semantic steering with the Fisher
information metric along a continuous diffusion trajectory.
It explicitly names LangFlow in both motivation and
approach. The highlighted correspondences with LangFlow
concern the continuous-generation setting and
perplexity-based evaluation.

\end{document}